\documentclass[a4paper,12pt]{amsart}
\usepackage[T1]{fontenc}
\usepackage[utf8]{inputenc}
\usepackage{mathtools}
\usepackage{amsthm,amssymb}
\usepackage[cal=boondoxo]{mathalfa}
\usepackage[pdfpagelabels]{hyperref}
\hypersetup{bookmarks,bookmarksopen,bookmarksdepth=2}
\usepackage[a4paper]{geometry}
\usepackage[inline]{enumitem}
\usepackage{csquotes}
\usepackage{stmaryrd}
\usepackage{microtype}
\usepackage{mathabx}
\usepackage{bbm}
\usepackage{booktabs}
\usepackage{xcolor}
\usepackage{nicematrix}
\usepackage{siunitx}
\usepackage{float}
\usepackage{tikz}
\usetikzlibrary{positioning}
\usepackage{hyperxmp}

\hypersetup{%
    pdftitle={On the loss landscape of the linear neural network: beyond critical points},
    pdfauthor={Antoine Lejay; Lionel Lenotre; Saïd Toubra},
    pdfdate={2026-11},
    pdfkeywords={linear neural network; critical point; thalweg; linear regression},
    pdflang={en},
    pdfmetalang={en},
}

\newtheorem{theorem}{Theorem}
\newtheorem{proposition}{Proposition}
\newtheorem{lemma}{Lemma}
\newtheorem{corollary}{Corollary}
\theoremstyle{definition}

\newtheorem{definition}{Definition}

\theoremstyle{remark}

\SetEnumitemKey{thm}{nolistsep,topsep=-\parskip,label={\textnormal{(\roman*)}}}
\SetEnumitemKey{compact}{noitemsep,topsep=-\parskip,leftmargin=0em,labelindent=2em}

\DeclarePairedDelimiterXPP{\normsup}[1]{}{\|}{\|}{_{\infty}}{#1}
\DeclarePairedDelimiterXPP{\norm}[1]{}{\|}{\|}{}{#1}
\DeclarePairedDelimiterXPP{\generalnorm}[2]{}{\|}{\|}{_{#1}}{#2}
\DeclarePairedDelimiterXPP{\normF}[1]{}{\|}{\|}{_{F}}{#1}

\DeclarePairedDelimiterXPP{\Frob}[1]{}{\langle}{\rangle}{_{F}}{#1} % Frobenius product
\DeclarePairedDelimiter{\abs}{|}{|}

\DeclarePairedDelimiter\Paren()

\DeclarePairedDelimiterXPP{\lieG}[1]{}{[}{]}{_{\prodG}}{#1}

\newcommand\given{\nonscript\:\delimsize\vert\nonscript\:\mathopen{}} 
\newcommand\SetSymbol[1][]{\nonscript\:#1\vert\nonscript\:\mathopen{}\allowbreak}
\DeclarePairedDelimiterX\Set[1]\{\}{%
  \renewcommand\given{\SetSymbol[\delimsize]}#1}
\DeclarePairedDelimiterX\Prb[1]{[}{]}{%
  \renewcommand\given{\SetSymbol[\delimsize]}#1}
  \DeclarePairedDelimiterXPP\Law[1]{\Lawname}{(}{)}{}{%
  \renewcommand\given{\SetSymbol[\delimsize]}#1}

\newcommand{\RR}{\mathbb{R}}

\DeclareMathOperator{\Rsq}{\mathrm{R}^2}
\DeclareMathOperator{\Rsquncent}{\mathrm{R}^2_{\mathrm{uncent}}}

\DeclareMathOperator*{\argmin}{arg\,min}

\newcommand{\cS}{\mathcal{S}}
\newcommand{\cSt}{\mathcal{S\!t}}

\newcommand{\cT}{\mathcal{T}}

\newcommand{\cN}{\mathcal{N}}
\newcommand{\cK}{\mathcal{K}}
\newcommand{\cU}{\mathcal{U}}
\newcommand{\cZ}{\mathcal{Z}}
\newcommand{\cL}{\mathcal{L}}
\newcommand{\cP}{\mathcal{P}}

\newcommand{\cE}{\mathcal{E}}

\newcommand{\cV}{\mathcal{V}}

\newcommand{\cI}{\mathcal{I}}
\newcommand{\cF}{\mathcal{F}}
\newcommand{\cH}{\mathcal{H}}
\newcommand{\cM}{\mathcal{M}}

\newcommand{\sZ}{\mathsf{Z}}

\newcommand{\dd}{\mathrm{d}}
\newcommand{\dD}{\mathrm{D}}

\DeclareMathOperator{\Tr}{Tr}

\DeclareMathOperator{\Id}{\mathsf{Id}}

\DeclareMathOperator{\Row}{Row}

\DeclareMathOperator{\Range}{Im}

\DeclareMathOperator{\rk}{\mathrm{rank}}

\newcommand{\sP}{\mathsf{P}}
\newcommand{\sR}{\mathsf{R}}

\newcommand{\sU}{\mathsf{U}}

\newcommand{\sw}{\mathsf{w}}

\newcommand{\cR}{\mathcal{R}}

\newcommand{\din}{d_{\mathrm{in}}}
\newcommand{\dout}{d_{\mathrm{out}}}
\newcommand{\dhid}{d_{\mathrm{hid}}}

\newcommand{\fF}{\mathfrak{F}}
\newcommand{\fE}{\mathfrak{E}}

\newcommand{\fG}{\mathfrak{G}}

\newcommand{\ff}{\mathfrak{f}}
\newcommand{\fh}{\mathfrak{h}}
\newcommand{\fk}{\mathfrak{k}}

\title[On the loss landscape]{On the loss landscape of the linear neural network: beyond critical points}

\author{Antoine Lejay \and Lionel Lenotre \and Saïd Toubra}
    \address{Lionel Lenotre: Université de Haute-Alsace, F-68100 Mulhouse, France.}
	\address{Antoine Lejay: Université de Lorraine, CNRS, Inria, IECL, F-54000 Nancy, France; ORCID: 0000-0003-0406-9550}
	\address{Saïd Toubra: Université de Lorraine, CNRS, Inria, IECL, F-54000 Nancy, France.}
\date{November 18, 2026}

\begin{document}

%% ABSTRACT
\begin{abstract}
	While the critical points of the loss landscape in linear neural networks have been extensively studied since the seminal work of Baldi \& Hornik in the late 1980s, a description of this landscape at an intermediate or global scale is still lacking. 
	Such a characterization is crucial, particularly as a substantial body of literature indicates that gradient descent circumvents critical points. The present article introduces an intermediate structure, termed the critical thalweg, which encompasses all critical points and exhibits key geometric properties. 
	Beyond providing a geometric proof of Baldi \& Hornik's results using simpler assumptions than the existing literature, this framework offers a global view of the loss landscape that can be easily visualized in a two-dimensional plane. 
	Empirical observations suggest that gradient descent is attracted toward the critical thalweg, along which the loss continues to decrease to its global minimum.
\end{abstract}

\maketitle

%%%%%%%%%%%%%%%%%%%%%%%%%%%%%%%%%%%%%%%%%%%%%%%%%%%%%%%%%%%%%%%%%%%%%%
%%%%%%%%%%%%%%%%%%%%%%%%%%%%%%%%%%%%%%%%%%%%%%%%%%%%%%%%%%%%%%%%%%%%%%

\section*{Introduction}

Demystifying the loss landscape of neural networks stands as a fundamental
challenge, even in the context of linear architectures. The interplay between
high-dimensional parameter spaces and non-convex geometries severely restricts
direct visualization and necessitates low-dimensional
projections~\cite{Li2018}.  Shedding light on critical points and their nature
is pivotal: it can drive the design of faster gradient-based
optimizers~\cite{Chaudhari2017,Ge2017b,Liu2025}, guide principled
initialization schemes~\cite{Arora2019a,Cohen2024}, and untangle complex
phenomena such as implicit regularization~\cite{Gidel2019}, feature
alignment~\cite{Ji2019}, or the effectiveness of residual
connections~\cite{Hardt2018}. Since the seminal work of Baldi \&
Hornik~\cite{Baldi1989}, a vast body of research has investigated the critical
points of linear neural
networks~\cite{Achour2024,Baldi2012,Dauphin2014,Ge2017a,Kawaguchi,Laurent,Nouiehed2022a,Trager2020,Zhang2020,Zhou2018}.
Despite an exponential growth in the number of critical points as
dimensionality increases, these landscapes exhibit a remarkably benign
structure devoid of spurious local
minima~\cite{Ge2017a,Ge2017b,Helmke1995,Lu2017,Nguyen2017}. Although the local
classification of these points is now well established, their global geometric
organization remains largely unexplored, as existing insights rely primarily on
elaborate frameworks from statistical physics~\cite{Choromanska2015}.

Optimizing general non-convex functions is NP-hard in the worst
case~\cite{Murty}; nevertheless, gradient-based algorithms routinely evade
pathological critical points in practice, even though rigorous convergence
guarantees remain confined to idealized regimes or restrictive
hypotheses~\cite{Arora2019b,Bah2022,Nguegnang2024,Panageas2019,Wendin2026}. The
ability of gradient methods to locate well-behaved minimizers has focused
intense interest on the geometric properties of reachable solutions. The
concept of \textquote{flat minima} (or wide valleys)~\cite{Hochreiter1997} has
notably gained traction as a candidate explanation for strong generalization,
under the intuition that regions where the loss remains low offer inherent
robustness against data perturbations~\cite{Han2025}. However, this perspective
remains actively debated. As highlighted by Dinh~\textit{et
al.}~\cite{Dinh2017}, measures of flatness are inherently sensitive to
reparameterization schemes that leave the end-to-end model invariant, meaning
sharp minima can be arbitrarily flattened without altering network
predictions~\cite{Fan2025,Frye2021,Petzka2021,Valavi2020}. This limitation
underscores the need for intrinsic geometric characterizations that remain
invariant under transformations of the parameter space, motivating a rigorous
examination of fundamental architectures such as linear neural networks.

The present article develops a framework centered on the notion of the
\emph{thalweg} to establish the absence of spurious local minima under minimal
assumptions (see Theorem~\ref{thm:critical_points}). Earlier studies,
notably~\cite{Kawaguchi} and subsequent extensions~\cite{Laurent,Lu2017},
established the absence of spurious local minima under restrictive hypotheses
such as full-rank target matrices~$Y$ or distinct singular values. Here,
\emph{only} the input data matrix~$X$ has full column rank. Unlike classical
proofs that rely on intricate case-by-case Hessian computations in the
parameter space, the thalweg approach constructs explicit ascending paths The
thalweg framework also provides an intrinsic, $2d$-dimensional graphical
representation of the optimization dynamics. 
The critical thalweg projects onto an explicit algebraic curve that contains
the image of all critical points of~the loss function. In this geometry, the uncentered
coefficient of determination is naturally interpreted and linked to the
spectral properties.
Numerical experiments corroborate these
theoretical findings, demonstrating that steepest gradient descent rapidly
enters in the critical thalweg and bypasses non-minimal critical points.

Several avenues remain open for future investigation. While the present
analysis characterizes the end-to-end matrix associated with critical points via orthogonal projectors, deriving
a closed-form analytical characterization of \emph{all} critical weight matrices 
remains an open problem. 
Our visualization framework motivates a quantitative dynamical
analysis of gradient flow and discrete gradient descent, aimed at establishing
convergence rates toward the critical thalweg and characterizing
escape times near saddle points. Finally, extending the
thalweg reduction and spectral projector methodology to regularized objectives, 
structured linear architectures, and mildly nonlinear networks represents a promising direction
toward bridging invariant geometric theory and practical deep learning
optimization.

\textbf{Outline.} Section~\ref{sec:thalweg} gives a foundational example that complete the introduction, 
formalizes the linear regression problem, introduces the ray-wise minimization framework, and
defines the critical and generalized thalwegs. Section~\ref{sec:rank} establishes the
equivalence between the critical points of the multi-layer objective~$\fG$ and
those of the loss~$\fF$ restricted to fixed-rank matrices.
Section~\ref{sec:critical} resolves the rank-constrained problem on 
orthogonal projectors, giving an alternative proof of Baldi \& Hornik result.
Section~\ref{sec:shape} exhibits paths with rank-increasing trajectories to
prove the absence of spurious local minima. Finally, Section~6 develops the
two-dimensional $(\rho, \theta)$-plane representation, establishes its
connection to the coefficient of determination~$\Rsq$, and presents numerical
experiments illustrating gradient descent
dynamics along the thalweg.

%%%%%%%%%%%%%%%%%%%%%%%%%%%%%%%%%%%%%%%%%%%%%%%%%%%%%%%%%%%%%%%%%%%%%%
%%%%%%%%%%%%%%%%%%%%%%%%%%%%%%%%%%%%%%%%%%%%%%%%%%%%%%%%%%%%%%%%%%%%%%

\section{Introducing the Thalweg}

\label{sec:thalweg}

%%%%%%%%%%%%%%%%%%%%%%%%%%%%%%%%%%%%%%%%%%%%%%%%%%%%%%%%%%%%%%%%%%%%%%

\subsection{A Foundational Example} \label{intro:sub:example}

An intuition for the loss landscape of deep linear networks can be found in the case of the simplest possible architecture: a two-layer linear network where the input, hidden, and output dimensions all equal one, and the weight matrices consequently reduce to real numbers, denoted by $v$ and $w$. 
Assuming a unit input $x=1$ and target $y=1$, and taking the quadratic loss, the problem consists of minimizing
\begin{equation*}
    \mathfrak{G}(v,w) := \frac{1}{2}(1 - vw)^2.
\end{equation*}
The minimum is achieved at $z_\star = 1$. 
The partial derivatives of $\fG$ evaluate as
\begin{equation} \label{eq:partial_derivative:example}
    \partial_v \fG(v,w) = -(1-vw)w \quad \text{and} \quad \partial_w \fG(v,w) = -(1-vw)v.
\end{equation}
A point $(v,w)$ is thus a critical point of $\fG$ if and only if $(v,w)=(0,0)$ or $vw=1$. 
At the origin, for any sufficiently small $\epsilon>0$, $\fG(\epsilon,\epsilon)=\frac{1}{2}(1-\epsilon^2)^2<\fG(0,0)$ and $\fG(\epsilon,-\epsilon)=\frac{1}{2}(1+\epsilon^2)^2>\fG(0,0)$, characterizing $(0,0)$ as a saddle point. 

\begin{figure}[htpb]
    \centering
    \includegraphics{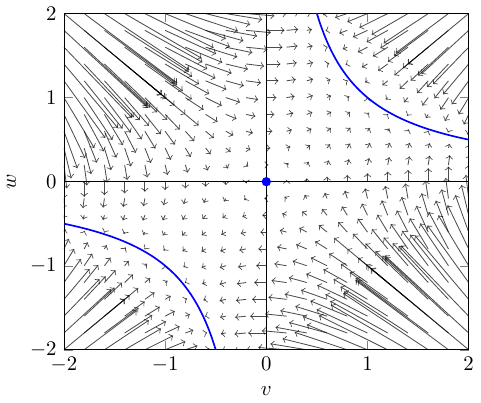}
    \caption{\label{fig:hyp} Vector field $-\nabla \mathfrak{G}(v,w)$. The hyperbolas form the locus of global minimizers of $\mathfrak{G}$, while the origin $(0,0)$ constitutes a strict saddle point.}
\end{figure}

Figure~\ref{fig:hyp} illustrates the vector field $-\nabla \fG$. 
The parameter space partitions into four distinct quadrants. 
For $vw<0$ (the lower-right and upper-left quadrants), the inequality $\fG(v,w) > \fG(0,0) = \frac{1}{2}$ holds. 
Along the axes $\Set{v=0}$ and $\Set{w=0}$, the function evaluates as $\fG(v,w)=\frac{1}{2}$.

Figure~\ref{fig:hyp} delineates level curves of $\fG$. 
The loss is in fact constantly equal to $\frac{1}{2}(1-z)^2$ for any $z\in\RR$ along $\cH_z := \Set{(v,w)\in\RR^2\given vw = z}$, which comprises two hyperbolic branches, determined by the respective signs of $v$ and $w$. 
For $z<0$, any point $(v,w)\in\cH_z$ strictly satisfies $\fG(v,w)>\frac{1}{2}$, while for $z>0$, the inequality $\fG(v,w)\geq \frac{1}{2}$ holds along $\cH_z$ if and only if $z\geq 2$. 
Consequently,
\begin{equation*}
	\Set{(v,w)\given \fG(v,w)<\fG(0,0)}=\bigcup_{0<z<2}\cH_z,
\end{equation*}
and $\fG$ vanishes entirely along $\cH_1$, which therefore encompasses all non-zero critical points, each acting as a global minimum.

Equation \eqref{eq:partial_derivative:example} reveals a cross-proportionality: the gradient with respect to $v$ is proportional to $w$, and vice versa. 
Gradient descent algorithms used in the learning process apply these derivatives to update the weights with a step size $\eta$:
\begin{equation*}
    v_{k+1} = v_k + \eta w_k(1-v_kw_k) \quad \text{and} \quad w_{k+1} = w_k + \eta v_k(1-v_kw_k).
\end{equation*}
The update rules share the identical error term $(1-v_kw_k)$. 
The squared difference $v^2 - w^2$ consequently evolve at the exact same absolute rate.

Eliminating the step-size artifacts of these discrete updates by passing to the continuous-time limit,
$v^2(t) - w^2(t)$ remains constant over the time as 
\begin{equation*}
    \frac{\mathrm{d}}{\mathrm{d}t} \left( v^2(t) - w^2(t) \right) = 2v(t)w(t)(1-v(t)w(t)) - 2v(t)w(t)(1-v(t)w(t)) = 0.
\end{equation*}
The gradient descent trajectories follow rigid hyperbolic paths instead of exploring the two-dimensional plane freely, crossing the loss contours $\mathcal{H}_z$ orthogonally as they move away from the saddle point at $(0,0)$ and toward the global minima on $\mathcal{H}_1$.
This geometric constraint justifies studying the product dynamics $z(t) = v(t)w(t)$ alone, bypassing the independent analysis of $v$ and $w$.

Introducing a scaling parameter $\lambda \ge 0$ for a fixed non-zero representation $z = vw$ yields the strictly convex restricted loss
\[
	\mathfrak{f}_z(\lambda) := \frac{1}{2}(1 - \lambda z)^2.
\] 
The unique minimum of $\mathfrak{f}_z$ is attained when 
\begin{equation} \label{eq:ortho:1d}
    \mathrm{d}\mathfrak{f}_z(\lambda) = -z(1 - \lambda z) = 0,
\end{equation}
leading to $\lambda z = z_\star = 1$ for $z \neq 0$.
This vanishing derivative encodes a geometric condition: the residual error $(1 - \lambda z)$ is orthogonal to $z$. 
Figure~\ref{fig:hyp} reveals that varying the parameter $\lambda$ geometrically corresponds to crossing the successive hyperbolic loss contours until landing on $\cH_1$. 
The scalar $z$ for which a fixed parameter $\lambda$ minimizes $\mathfrak{f}_z$ defines a specific one-dimensional locus. 
This set forms a collection of minima across the loss landscape, acting as the exact mathematical equivalent of a geological \emph{thalweg}, which connects the lowest points of a valley floor.
It satisfies the orthogonality expressed in Equation~\eqref{eq:ortho:1d} and isolates the critical points at $\lambda = 1$, endowing the critical set with an algebraic structure.

%%%%%%%%%%%%%%%%%%%%%%%%%%%%%%%%%%%%%%%%%%%%%%%%%%%%%%%%%%%%%%%%%%%%%%

\subsection{Formal Definition and Properties}

Let $\mathcal{M}(r, s)$ denote the space of real matrices with $r$ rows and $s$ columns. A $p$-layer linear neural network maps an input matrix $X \in \mathcal{M}(n, \din)$ to an output $Y \in \mathcal{M}(n, \dout)$ through a predictor
\begin{equation*}
    \widehat{Y} = X V_1 \dotsb V_p \quad \text{with } V_k \in \mathcal{M}(d_{k-1}, d_k),
\end{equation*}
where $d_0 = \din$ and $d_p = \dout$. 
The matrices $(V_1, \dots, V_p)$ are the network weights. 

Taking the quadratic loss frames the learning process as the minimization of
\begin{equation*} 
    %\label{eq:def:G}
    \mathfrak{G}(V_1, \dots, V_p) := \frac{1}{2}\normF{Y - X V_1 \dotsb V_p}^2,
\end{equation*}
where $\normF{\cdot}$ denotes the Frobenius norm, induced by the inner product 
\begin{equation*}	
    \Frob{A,B}
	 = \Tr(A^\top B), \quad \forall A, B \in \mathcal M(n,\dout).
\end{equation*}
Once the parameter space is denoted by 
\begin{equation*}
\cM(d_0,\dotsc,d_p) := \cM(d_0,d_1)\times\dotsb\times\cM(d_{p-1},d_p),
\end{equation*}
such a network can be reparametrized via the mapping 
\begin{equation*}
    \Psi_p:(V_1,\dotsc,V_p)\in \cM(d_0,\dotsc,d_p)\longmapsto V_1\dotsb V_p\in\cM(\din,\dout),
\end{equation*}
yielding the function $\mathfrak{F} \colon \cM(d_0, d_p) \to \mathbb{R}$ defined by
\begin{equation} 
    \label{eq:def:F}
    \mathfrak{F}(Z) := \frac{1}{2}\normF{Y - XZ}^2, \quad \forall Z \in \cM(\din, \dout).
\end{equation}

Since $X$ has full column rank, $\mathfrak{F}$ is strictly convex and its unique minimizer is given by the ordinary least squares solution
\begin{equation}
    \label{eq:value:Zstar}
    Z_{\star} = \Sigma_{X,X}^{-1}\Sigma_{X,Y}, \quad \text{where } \Sigma_{M,N} := M^\top N \text{ for compatible } M, N.
\end{equation}
The global minimum value evaluates as
\begin{equation*} 
    \mathfrak{F}(Z_{\star}) = \frac{1}{2}\normF{Y}^2 - \frac{1}{2}\Tr(\Sigma_{Y,X} \Sigma_{X,X}^{-1} \Sigma_{X,Y}),
\end{equation*}
and is achieved by the network whenever its bottleneck dimension $\underline{d}:=d_0\wedge \dotsb \wedge d_k$ satisfies $\underline{d} \ge \rk(Z_{\star})$.
Otherwise, $Z_{\star}$ cannot be reached, restricting the problem to the search for some solution $Z_\dag$ of $\mathfrak{F}$ with rank at most $\underline{d}$, for which the loss decomposes as
\begin{equation*}
    \mathfrak{F}(Z_\dag) = \mathfrak{F}(Z_{\star}) + \frac{1}{2} \normF{X(Z_\dag - Z_{\star})}^2.
\end{equation*}
When $X$ is the identity matrix, finding $Z_\dag$ is equivalent to finding the best low-rank approximation of $Z_\star$, which is given by the Eckart--Young theorem~\cite{Eckart}.

Regardless of the rank of $Z \in \cM(\din, \dout)$, the mapping $B \mapsto \fF(BZ)$ is convex.
The set of minimizers of this mapping over $\cM(\din, \din)$ is therefore non-empty, a property that ensures the well-posedness of an object called the \emph{Thalweg}
which is the main tool of this paper.

\begin{definition}[Thalweg]
    \label{def:thalweg}
    For a given matrix $A \in \cM(\din, \din)$, the \emph{thalweg} associated with $A$ is defined as
    \begin{equation*}
        \cT(A) := \Set*{Z \in \cM(\din, \dout) \given A \in \argmin_{B \in \cM(\din, \din)} \fF(BZ)}.
    \end{equation*}
\end{definition}

Matrices in the thalweg enjoy fundamental properties stated in Proposition~\ref{prop:thalweg}.
The orthogonality relation \eqref{eq:thalweg:3} yields Corollary~\ref{cor:thalweg}, showing that minimizing the loss over the set
\begin{equation*}
    \cR(Z) := \Set*{ Z' \in \cM(\din, \dout) \given \Row(Z') \subseteq \Row(Z) }
\end{equation*}
reduces to an orthogonal projection of $Z_\star$ onto the row space of $Z$. This also states that every matrix belongs to at least one thalweg.

\begin{proposition} 
    \label{prop:thalweg}
	Let $A \in \cM(\din, \din)$ and $Z \in \cM(\din, \dout)$. The condition $Z \in \cT(A)$ is equivalent to stating
	\begin{equation} \label{eq:thalweg:1}
	    \dD\fF(AZ) \cdot (BZ) = -\Frob{Y - XAZ, XBZ} = 0, \quad \forall B \in \cM(\din, \din),
	\end{equation}
	which is equivalent to the difference $\fF(Z') - \fF(AZ)$ being purely quadratic for any~$Z'$ whose row space is included in that of~$Z$, that is,
    \begin{equation}
        \label{eq:thalweg:2}
        \fF(BZ) = \fF(AZ) + \frac{1}{2}\normF{X(B-A)Z}^2, \quad \forall B \in \cM(\din, \din).
    \end{equation}
   	These two conditions are equivalent to the orthogonality relation
    \begin{equation}
        \label{eq:thalweg:3}
        (X^\top Y - X^\top X A Z)Z^\top = (\Sigma_{X,Y} - \Sigma_{X,X} A Z)Z^\top = 0.
    \end{equation}
\end{proposition}

\begin{proof}
	By the convexity of $B \mapsto \fF(BZ)$, $Z \in \cT(A)$ if and only if the differential of~$\fF$ at $AZ$ along any direction $K$ given by
	    $\dD \fF(A) \cdot K = -\Frob{Y-XA, XK}$ vanishes.
	Evaluating  $\dD \fF(A) \cdot K$ in the direction $K=BZ$ directly yields Equation~\eqref{eq:thalweg:1}.
	
	The quadratic function $\fF$ expands as
	\begin{equation*}
	    \fF(BZ) = \fF(AZ) - \Frob{Y-XAZ, X(B-A)Z} + \frac{1}{2}\normF{X(B-A)Z}^2.
	\end{equation*}
	Equation~\eqref{eq:thalweg:1} holds for any arbitrary matrix $B$, applying it to the difference $(B-A)$ cancels the cross-term, establishing~\eqref{eq:thalweg:2}.
    
	By the cyclic property of the trace, the condition in Equation~\eqref{eq:thalweg:1} rewrites as
	\begin{equation*}
	    \Frob{X^\top(Y - XAZ)Z^\top, B} = 0,
	\end{equation*}
	which implies the orthogonality relation~\eqref{eq:thalweg:3} since this equality must hold for all $B \in \cM(\din, \din)$.
\end{proof}

\begin{corollary} 
    \label{cor:thalweg}
    Every matrix $Z \in \cM(\din, \dout)$ belongs to at least one thalweg, and for all $Z' \in \cR(Z)$,
    \begin{equation*}
        \fF(Z') \geq \fF(Z_\star Z^+Z),
    \end{equation*}
    where $Z^+Z$ denotes the orthogonal projector onto the row space of $Z$. 
    Moreover, the minimum over $\cR(Z)$ is uniquely reached at $Z' = Z_\star Z^+Z$.
\end{corollary}

\begin{proof}
	Since $\Sigma_{X,X}$ is invertible, Equation~\eqref{eq:thalweg:3} rewrites
	\begin{equation*}
	    \Sigma_{X,X}^{-1}\Sigma_{X,Y}Z^\top - AZZ^\top = Z_{\star}Z^\top - AZZ^\top =0.
	\end{equation*}
	Denoting by $(ZZ^\top)^+$ the Moore-Penrose inverse of $ZZ^\top$, the product $(ZZ^\top)^+(ZZ^\top)$ corresponds to the orthogonal projector $ZZ^+$. 
	Hence, from the symmetry of this projector and the fundamental properties of the generalized inverse,
	\begin{equation*}
	    Z^\top (ZZ^\top)^+(ZZ^\top) = Z^\top (ZZ^+)^\top = (ZZ^+Z)^\top = Z^\top.
	\end{equation*}
	According to \cite[Sect.~3]{Rao}, the equation $AZZ^\top = Z_{\star}Z^\top$ admits a solution $A$ if and only if $Z_{\star}Z^\top(ZZ^\top)^+(ZZ^\top) = Z_{\star}Z^\top$.
	Since this condition trivially holds from the above identity, applying the standard properties $Z^\top(ZZ^\top)^+ = Z^+$ and $(ZZ^\top)(ZZ^\top)^+ = ZZ^+$ simplifies any solution $A$ to
	\begin{equation*}
	    A = Z_{\star}Z^+ + B(\Id - ZZ^+) \quad \text{for any matrix } B.
	\end{equation*}

	Right-multiplying any such $A$ by $Z$ yields $AZ = Z_{\star}Z^+Z$. 
	Since any $Z' \in \cR(Z)$ satisfies $Z' = (Z'Z^+)Z$, applying Equation~\eqref{eq:thalweg:2} yields $\fF(Z') \geq \fF(AZ) = \fF(Z_{\star}Z^+Z)$, and $Z_{\star}Z^+Z$ achieves the minimum over $\cR(Z)$ as an element of this set.
	Finally, the strict convexity of $\fF$ over $\cR(Z)$ ensures the uniqueness of this minimum.
\end{proof}

The thalweg admits a geometric interpretation that mirrors the topographical concept of a valley floor.
Figure~\ref{fig:thalweg} shows black curves representing the loss evaluated over the linear subspaces $\cR(Z)$. 
Beneath them, the red line acts as the valley floor connecting restricted minima, like the projected point $Z_\star Z^+Z$, to the global minimum $Z_\star$.
While this low-dimensional figure captures the essence of the Thalweg, the high-dimensional setting of the actual problem implies that these paths might follow far more unexpected routes.

A natural generalization of the thalweg is obtained by restricting the direction of perturbations to a submanifold $\cN$ of $\cM(\din,\din)$:
\begin{equation}
    \label{eq:thalweg-generalized}
    \cT_{\cN}(A) := \Set*{Z\in\cM(\din,\dout) \given A \in \argmin_{B\in\cN} \ff_Z(B)}.
\end{equation}
This restriction contains the unconstrained thalweg : $\cT(A) \subset \cT_{\cN}(A)$. 
The orthogonality condition from Lemma~\ref{prop:thalweg} translates to 
\begin{equation*}
    \Frob{Y-X AZ,XBZ}=0  \text{ for any } B\in\cN.
\end{equation*}

The thalweg $\cT(\Id)$ is called the \emph{critical thalweg} as it captures the critical points of the loss. 
Proposition~\ref{prop:critical_thalweg} highlights the importance of this thalweg by showing that both restricted and partial critical points necessarily belong to it.

\begin{proposition}
    \label{prop:critical_thalweg}
    If $Z$ is a critical point of $\fF$ restricted to the manifold $\cM_{r}(\din,\dout)$ of matrices of rank $r$, then $Z$ belongs to the thalweg $\cT(\Id)$. 
   	Furthermore, if $(V_1,\dotsc,V_p)$ is a partial critical point of $\mathfrak{G}$ with respect to the first factor $V_1$, the product $Z = V_1\dotsb V_p$ also belongs to $\cT(\Id)$.
\end{proposition}

\begin{proof}
	Let $Z$ be a critical point of $\fF$ restricted to $\cM_r(\din,\dout)$, which is a manifold as a submanifold of the ambient matrix space \cite[Lemma~1]{Baldi2012}.
	For any matrix $B$, the path $\sZ(t) = (\Id+tB)Z$ preserves the rank of $Z$ for sufficiently small $t$ as $\Id+tB$ is invertible.
	This path remains within $\cM_r(\din,\dout)$ and yields the tangent vector $\dd_{t}{\sZ}(0) = BZ$, along which the differential of $\fF$ must vanish, giving
	\begin{equation*}
	    \dD\fF(Z)\cdot (BZ) = \Frob{Y-XZ,XBZ} = \Frob{(X^\top Y-X^\top X Z)Z^\top,B} = 0.
	\end{equation*}
	As this holds for any matrix $B$, $(X^\top Y-X^\top XZ)Z^\top = 0$, which by Proposition~\ref{prop:thalweg} implies $Z \in \cT(\Id)$.

	Let $Z = V_1\dotsb V_p$. Since $\nabla \ff_Z(\Id) = 0$ when $Z = 0$, it suffices to consider the case $Z \neq 0$.
    For any arbitrary matrix $B \in \cM(\din,\din)$, 
    \begin{equation*}
        \fG(B V_1, V_2, \dotsc, V_p) = \fF(B V_1 V_2 \dotsb V_p) = \fF(B Z) = \ff_Z(B).
    \end{equation*}
    By the chain rule, 
    \begin{equation*}
        \nabla \ff_Z(\Id) = \nabla_{V_1} \fG(V_1,\dotsc,V_p) \cdot B V_1.
    \end{equation*}
    Since $(V_1,\dotsc,V_p)$ is a critical point with respect to $V_1$, $\nabla_{V_1} \fG(V_1,\dotsc,V_p) = 0$, 
    which implies $\nabla \ff_Z(\Id) = 0$. 
    Because $\ff_Z$ is a convex quadratic function, any critical point is a global minimizer and $Z \in \cT(\Id)$.
\end{proof}

%%%%%%%%%%%%%%%%%%%%%%%%%%%%%%%%%%%%%%%%%%%%%%%%%%%%%%%%%%%%%%%%%%%%%%
%%%%%%%%%%%%%%%%%%%%%%%%%%%%%%%%%%%%%%%%%%%%%%%%%%%%%%%%%%%%%%%%%%%%%%

\section{Reduction to a Rank Constraint Problem}

\label{sec:rank}

The product structure of deep networks obscures a direct analysis of the critical points of $\fG$ in the multi-layer parameter space $\cM(d_0,\dotsc,d_p)$. 
The difficulty disappears once the layers collapse into a single matrix $Z = \Psi_p(V_1,\dotsc,V_p)$, which transfers the problem to the loss~$\fF$ subject to the architectural rank bottleneck $\rk(Z) \le \underline{d}$. 
The construction of a smooth lifting map that embeds any rank-$r$ factorization back into the layer space establishes an exact correspondence between the critical points of~$\fG$ and those of~$\fF$ restricted to fixed-rank manifolds $\cM_r(\din,\dout)$. 
The optimization problem reduces to a rank-constrained matrix formulation, whose critical set can then be characterized algebraically.

\subsection{From layers to a single matrix.}

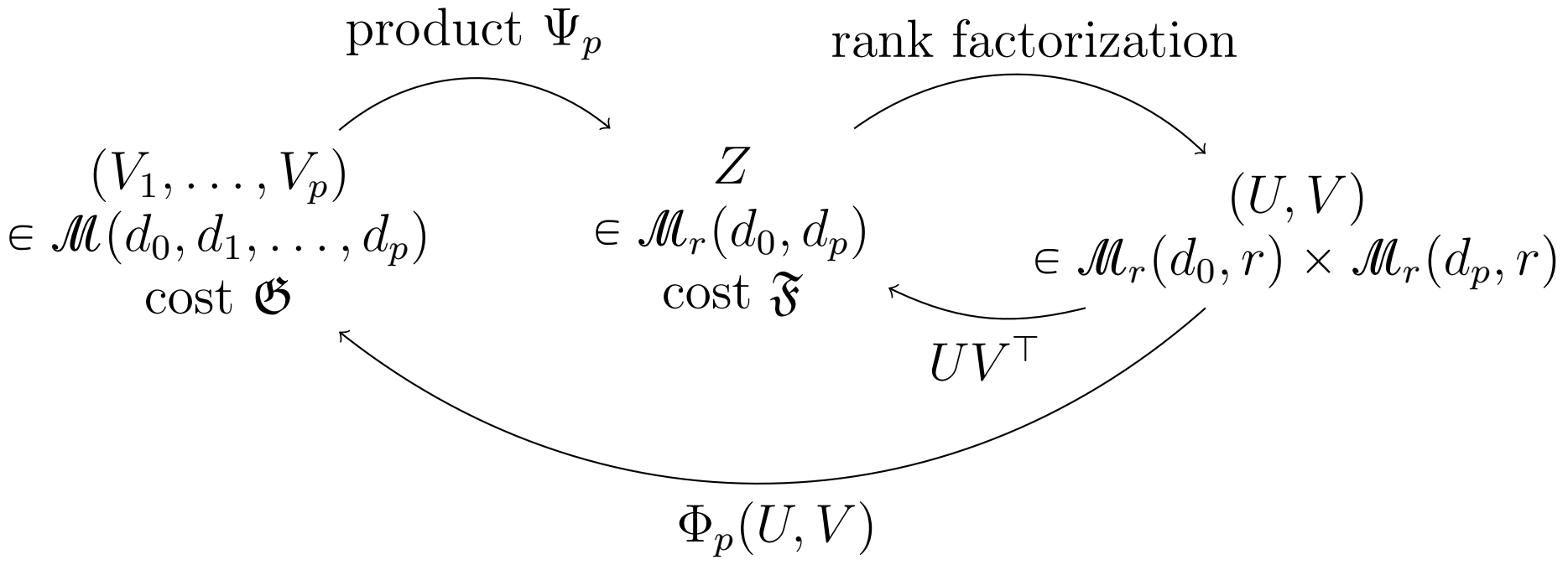
\begin{figure}
	\centering
	\begin{tikzpicture}
		\node[minimum width=2cm,align=center] (L) {$(V_1,\dotsc,V_p)$\\$\in \cM(d_0,d_1,\dotsc,d_p)$\\ cost $\fG$};
		\node (M) [right=1cm of L,minimum width=2cm, align=center] {$Z$\\$\in\cM_r(d_0,d_p)$\\ cost $\fF$};
		\node (R) [right=1cm of M,minimum width=2cm, align=center] {$(U,V)$\\$\in\cM_r(d_0,r)\times\cM_r(d_p,r)$};
		
		\draw[->, bend left=40] (L) to node[above] {product $\Psi_p$} (M);
		\draw[->, bend left=40] (R) to node[below] {$\Phi_p(U,V)$} (L);
		\draw[->, bend left=40] (M) to node[above] {rank factorization} (R);
		\draw[->, bend left=20] (R) to node[below] {$UV^\top$} (M);
	\end{tikzpicture}
	\caption{\label{fig:rank_layer} Equivalence between rank constrained matrices and space of layers.}
\end{figure}

Every weight tuple $(V_1,\dotsc,V_p)$ induces a product matrix $Z = \Psi_p(V_1,\dotsc,V_p)$ whose rank is bounded by the minimal layer dimension~$\underline{d}$. 
Figure~\ref{fig:rank_layer} illustrates the resulting interplay between the multi-layer parameter space, rank-constrained matrices, and rank factorizations. 
Proposition~\ref{prop:critical_equivalence} formalizes this correspondence for critical points, building on the smooth lifting map constructed in Lemma~\ref{lem:Psi}.

\begin{lemma}
    \label{lem:Psi}
    For $r \leq \underline{d}$, a smooth map $\Phi_p : \cM(d_0,r) \times \cM(d_p,r) \to \cM(d_0, \dots, d_p)$ satisfying the following for any $(U,V) \in \cM_r(d_0,r) \times \cM_r(d_p,r)$ can be defined:
    \begin{equation*}
	%\label{eq:Psi:1}
        \rk(\Psi_p(\Phi_p(U,V))) = \rk(UV^\top) \quad \text{and} \quad \Psi_p(\Phi_p(U,V)) = UV^\top.
    \end{equation*}
\end{lemma}

\begin{proof}
	Let $(U,W)\in \cM(d_0,r)\times\cM(d_p,r)$.
	Setting the boundary and intermediate factors ($1 < k < p$) to
	\begin{gather*}
	    V_1 = \begin{bmatrix} U & \mathbf{0} \end{bmatrix}\in\cM(d_0,d_1), \quad
	V_p = \begin{bmatrix} W^\top\\ \mathbf{0} \end{bmatrix}\in\cM(d_{p-1},d_p),
	\\
	\text{and }
	V_k = \begin{bmatrix} \Id_r & \mathbf{0} \\ \mathbf{0} & \mathbf{0} \end{bmatrix}\in\cM(d_{k-1},d_k), 
	\end{gather*}
	where $\Id_r$ denotes the $r \times r$ identity matrix, yields 
	$V_1 \dotsb V_p = U \Id_r \dotsb \Id_r V^\top =  UW^\top$
	by block multiplication. The function $\Phi_p$ is then defined as 
	$\Phi_p(U,W)=(V_1,\dotsc,V_p)$, and clearly is a smooth map. As $\Psi_p\circ \Phi_p(U,V) = UV^\top$, the rank equality holds.  
\end{proof}

\begin{proposition}
    \label{prop:critical_equivalence}
    If $(V_1, \dotsc, V_p)$ is a critical point of~$\mathfrak{G}$, then $Z = V_1 \dotsb V_p$ has rank $r \le \underline{d}$ and is a
    critical point of $\mathfrak{F}_{|\cM_r(\din,\dout)}$.
    Conversely, for any $r \le \underline{d}$, every critical point $Z$ of $\mathfrak{F}_{|\cM_r(\din,\dout)}$ can be lifted to a critical point $(V_1, \dotsc, V_p)$ 
    of~$\mathfrak{G}$ satisfying $V_1 \dotsb V_p = Z$.
\end{proposition}

\begin{proof}
	The gradient of $\fF$ is $\nabla \fF(Z) = \Sigma_{X,X}Z - \Sigma_{X,Y}$. 
	At $Z=UW^\top \in \cM_r(\din,\dout)$, the tangent space is formed by matrices of the form $UH^\top + K^\top W^\top$. 
	The condition $\dD{\fF(Z)} \cdot \dd{Z} = 0$ thus reads
	\begin{equation*}
	    \Frob{U^\top\nabla \fF(Z),H^\top} + \Frob{\nabla \fF(Z)W,K^\top} = 0, \quad \text{for all } H, K,
	\end{equation*}
	and $Z$ is a critical point of $\fF_{|\cM_r(\din,\dout)}$ if and only if 
	    \begin{equation} \label{eq:gradF:1}
			\nabla \fF(Z)W = 0 \text{ and } U^\top \nabla \fF(Z)=0.
	    \end{equation}
	    
	Let $Z$ be a critical point of $\fF_{|\cM_r(\din,\dout)}$ with rank factorization $Z=UW^\top$. 
	Defining $(V_1,\dotsc,V_p) = \Phi_p(U,W)$, for any $k \in \{2,\dotsc,p-1\}$,
	\begin{equation*}
	    \prod_{i=1}^{k-1} V_i = V_1 = \begin{bmatrix} U & \mathbf{0} \end{bmatrix} 
	    \quad \text{and} \quad
	    \prod_{i=k+1}^{p} V_i = V_p = \begin{bmatrix} W^\top \\ \mathbf{0} \end{bmatrix}.
	\end{equation*}
	Hence,
	\begin{equation} \label{eq:gradG}
	    \nabla_{V_k}\fG(V_1,\dotsc,V_p) = 
	    \begin{cases} 
	        \begin{bmatrix} U^\top \nabla \fF(Z) \\ \mathbf{0} \end{bmatrix} 
	        & \text{if } k=1, \\[3ex]
	        \begin{bmatrix} U^\top\nabla \fF(Z)W & \mathbf{0} \\ \mathbf{0} & \mathbf{0} \end{bmatrix} 
	        & \text{if } k=2,\dotsc,p-1, \\[3ex]
	        \begin{bmatrix} \nabla \fF(Z) W & \mathbf{0} \end{bmatrix} 
	        & \text{if } k=p.
	    \end{cases}
	\end{equation}
	Using the chain rule, for all $k \in \{1,\dotsc,p\}$,
	\begin{multline*}
	    \dD_{V_k}\fG(V_1,\dotsc,V_p)(H_k) 
	    = -\Frob{Y - X V_1\dotsb V_p, \, X V_1\dotsb V_{k-1} H_k V_{k+1}\dotsb V_p} \\
	    = \Frob{\nabla_{V_k}\fG(V_1,\dotsc,V_p), \, H_k},
	\end{multline*}
	which yields
	\begin{equation*} 
	    \nabla_{V_k}\fG(V_1,\dotsc,V_p) = (V_1\dotsb V_{k-1})^\top \nabla\fF(Z) (V_{k+1}\dotsb V_p)^\top.
	\end{equation*}
	Since Equation~\eqref{eq:gradF:1} holds at the critical point $Z$, it follows from Equation~\eqref{eq:gradG} that $\nabla_{V_k}\fG(V_1,\dotsc,V_p) = \mathbf{0}$ for all $k$. 
	$(V_1,\dotsc,V_p)$ is thus a critical point of $\fG$.

	Assuming $(V_1, \dotsc, V_p)$ is a critical point of $\fG$ and writing 
	\begin{equation*}
	    Z = U W^\top = V_1 (V_2 \dotsb V_p) = (V_1 \dotsb V_{p-1}) V_p,
	\end{equation*}
	$r = \rk(Z) \le \underline{d}$, and the row and column spaces of $Z$ are respectively contained in those of the right factor $(V_2 \dotsb V_p)$ and the left factor $(V_1 \dotsb V_{p-1})$. 
	There exist two matrices $M_W \in \mathbb{R}^{d_1 \times r}$ and $M_U \in \mathbb{R}^{d_{p-1} \times r}$ such that $W = (V_2 \dotsb V_p)^\top M_W$ and $U = (V_1 \dotsb V_{p-1}) M_U$.
	Applying the chain rule, the vanishing gradients of $\mathfrak{G}$ yield
    \begin{align*}
        \nabla \mathfrak{F}(Z) V &= \nabla \mathfrak{F}(Z) (V_2 \dotsb V_p)^\top M_V = \nabla_{V_1} \mathfrak{G}(V_1,\dotsc,V_p) \, M_W = 0, \\
        U^\top \nabla \mathfrak{F}(Z) &= M_U^\top (V_1 \dotsb V_{p-1})^\top \nabla \mathfrak{F}(Z) = M_U^\top \nabla_{V_p} \mathfrak{G}(V_1,\dotsc,V_p) = 0,
    \end{align*}
    and from \eqref{eq:gradF:1}, $Z = V_1 \dotsb V_p$ is a critical point of $\mathfrak{F}_{|\cM_r(\din,\dout)}$.
\end{proof}

%%%%%%%%%%%%%%%%%%%%%%%%%%%%%%%%%%%%%%%%%%%%%%%%%%%%%%%%%%%%%%%%%%%%%%

\subsection{Characterization of the Critical Thalweg}

\label{sec:thalweg:critical}

According to Proposition~\ref{prop:critical_equivalence}, critical points $(V_1,\dotsc,V_p)$ of~$\fG$ are in correspondence with critical points $Z$ of~$\fF$ restricted to matrices of rank at most $\underline{d}$.
By Proposition~\ref{prop:critical_thalweg}, the critical points belong to the critical thalweg~$\cT(\Id)$.
Theorem~\ref{thm:thalweg_characterization} and Corollary~\ref{cor:FZ_expansion} below parametrize $\cT(\Id)$ using the set of orthogonal projectors
\begin{equation*}
    \cP := \bigcup_{r=0}^{\underline{d}}\cP_r \quad \text{with} \quad
    \cP_r := \Set*{P \given P^\top=P,\ P^2=P,\ \rk(P)=r}.
\end{equation*}
Defining $\Sigma := \Sigma_{Y,X} \Sigma_{X,X}^{-1} \Sigma_{X,Y}$, the cost function restricted to the set $\Set{Z_{\star} P\given P\in\cP}$ becomes, by Corollary~\ref{cor:FZ_expansion},
\begin{equation}
    \label{eq:thalweg:7}
     \mathfrak{F}(Z_\star P) = \frac{1}{2}\normF{Y}^2 - \frac{1}{2}\mathfrak{h}(P), \quad \text{where } \mathfrak{h}(P) := \Frob{\Sigma,P} = \Tr(\Sigma P).
\end{equation}
The minimal value $\mathfrak{F}(Z_{\star})=\frac{1}{2}\normF{Y}^2-\frac{1}{2}\Tr(\Sigma)$ is then recovered by setting $P = \Id$.

\begin{theorem}
    \label{thm:thalweg_characterization}
    Let $A \in \mathcal (\din,\din)$ be invertible.
    A matrix $Z \in \cM(\din,\dout)$ belongs to the thalweg $\cT(A)$ if and only if it can be expressed as
    \begin{equation*}
        Z = A^{-1} Z_\star P,
    \end{equation*}
    where $P=Z^+Z \in \mathcal M(\dout,\dout)$ is the orthogonal projector onto the row space of $Z$ 
    (in particular, $\rk(P)=\rk(Z)$).
\end{theorem}

\begin{proof}
    Assume $Z = A^{-1}Z_\star P$, where $P$ is an orthogonal projector. 
    Recalling that $Z_\star=\Sigma_{X,X}^{-1}\Sigma_{X,Y}$, the right-hand side of Equation~\eqref{eq:thalweg:2} evaluates as
    \begin{align*}
	\Sigma_{X,X}AZZ^\top 
	&= \Sigma_{X,X} A A^{-1}\Sigma_{X,X}^{-1}\Sigma_{X,Y} PP^\top \Sigma_{Y,X}\Sigma_{X,X}^{-1}(A^{-1})^\top \\
	&= \Sigma_{X,Y} PP^\top \Sigma_{Y,X}\Sigma_{X,X}^{-1}(A^{-1})^\top.
    \end{align*}
    Since $P$ is an orthogonal projector,	
    \[
	\Sigma_{X,X}AZZ^\top = \Sigma_{X,Y} P \Sigma_{Y,X}\Sigma_{X,X}^{-1}(A^{-1})^\top = \Sigma_{X,Y}Z^\top.
    \]
    The condition in Equation~\eqref{eq:thalweg:2} thus holds, and $ZP\in\cT(A)$.

    Conversely, assume $Z \in \cT(A)$, meaning that Equation~\eqref{eq:thalweg:2} holds. 
    Since $X$ has full rank (so $\Sigma_{X,X}$ is invertible) and $A$ is invertible, left-multiplying both sides by $A^{-1}\Sigma_{X,X}^{-1}$ and recalling that $Z_\star = \Sigma_{X,X}^{-1}\Sigma_{X,Y}$ directly yields $A^{-1}Z_\star Z^\top = ZZ^\top$. 
    This simplifies to
    \begin{equation}
	\label{eq:proof:1}
	(A^{-1}Z_\star - Z) Z^\top = 0.
    \end{equation}
    Set $P=Z^+Z$, where $Z^+$ is the Moore-Penrose inverse of $Z$. 
    The matrix $P$ is the orthogonal projector onto the row space of $Z$ \cite[Dm 3, Proposition~3.5]{Sibuya}. 
    By definition of a generalized inverse, $ZP=ZZ^+Z=Z$, which implies
    \begin{equation*}
	A^{-1} Z_\star P - Z = A^{-1} Z_\star P - ZP = (A^{-1}Z_\star -Z)P.
    \end{equation*}
    Furthermore, since $Z^+=Z^\top(ZZ^\top)^+$ \cite[DMP 9, Proposition~4.8]{Sibuya} and $P=Z^+Z$, it follows from Equation~\eqref{eq:proof:1} that 
    \begin{equation*}
	A^{-1} Z_\star P - Z 
	= (A^{-1}Z_\star-Z)Z^\top (ZZ^\top)^+ Z = 0.
    \end{equation*}
    Hence, $Z=A^{-1} Z_{\star} P$ with $P=Z^+Z$.
\end{proof}

\begin{corollary}
\label{cor:FZ_expansion}
Let $Z \in \cT(\lambda \Id)$ with $\lambda \neq 0$, and let $P \in \mathcal{M}(\dout,\dout)$ be the orthogonal projector such that $Z = \lambda^{-1} Z_\star P$. Then,
\begin{equation}
    \label{eq:prop:FZ_expansion:1}
    \mathfrak{F}(Z) = \frac{1}{2}\normF{Y}^2 + \frac{1 - 2\lambda}{2 \lambda^2}\Frob{\Sigma,P}.
\end{equation}
\end{corollary}

\begin{proof}
	Substituting $Z = \lambda^{-1} Z_\star P$ and recalling that $Z_\star = \Sigma_{X,X}^{-1} \Sigma_{X,Y}$, the expressions for $\Frob{Y,XZ}$ and $\normF{XZ}^2$ evaluate respectively to
	\begin{align*}
	    \Frob{Y,XZ} 
	    &= \Frob{X^\top Y, \lambda^{-1} Z_\star P} 
	    = \lambda^{-1} \Frob{\Sigma, P}, \\
	    \normF{XZ}^2 
	    &= \lambda^{-2} \Frob{\Sigma_{X,X} \Sigma_{X,X}^{-1} \Sigma_{X,Y}, \Sigma_{X,X}^{-1} \Sigma_{X,Y} P P^\top} = \lambda^{-2} \Frob{\Sigma, P}.
	\end{align*}
	Substituting these evaluated terms back into the expansion
	\begin{equation*}
	    \mathfrak{F}(Z) = \frac{1}{2}\normF{Y}^2 - \Frob{Y,XZ} + \frac{1}{2}\normF{XZ}^2,
	\end{equation*}
	and factoring out $\Frob{\Sigma,P}$ yields Equation~\eqref{eq:prop:FZ_expansion:1}.
\end{proof}

%%%%%%%%%%%%%%%%%%%%%%%%%%%%%%%%%%%%%%%%%%%%%%%%%%%%%%%%%%%%%%%%%%%%%%

%%%%%%%%%%%%%%%%%%%%%%%%%%%%%%%%%%%%%%%%%%%%%%%%%%%%%%%%%%%%%%%%%%%%%%

\section{Identifying critical points through Projectors}

\label{sec:critical}

\subsection{Solving the Rank-Constrained Problem}

Directly characterizing the critical points of $\fG$ is complicated.
Restricting the end-to-end matrix $Z = V_1\dotsb V_p$ to the critical
thalweg~$\cT(\Id)$ resolves this difficulty: $\cT(\Id)$ is parametrized by orthogonal projectors.
Transforming the minimization of the loss~$\fF$ into the maximization of
the linear trace objective $\fh(P) = \Tr(\Sigma P)$.

Let $\lambda_1 > \dotsb > \lambda_q \geq 0$ be the distinct eigenvalues of $\Sigma$, where each $\lambda_k$ has multiplicity $m_k$.
A hierarchy $\cL$ of integer vectors $\boldsymbol{\nu} = (\nu_1, \dots, \nu_q)$, called \emph{signatures}, can be constructed over these eigenvalues,
where each integer $0 \le \nu_k \le m_k$ counts the selections of $\lambda_k$ up to its algebraic multiplicity $m_k$.
For two signatures~$\boldsymbol{\nu}$ and~$\boldsymbol{\gamma}$, whose sizes $\abs{\boldsymbol{\nu}}$ and $\abs{\boldsymbol{\gamma}}$ are the sums of their integers, a total preorder is defined by their energy levels:
\begin{equation*}
    \boldsymbol{\nu} \preceq \boldsymbol{\gamma} \quad \text{if} \quad \fE(\boldsymbol{\nu}) \leq \fE(\boldsymbol{\gamma}), \quad \text{where} \quad \fE(\boldsymbol{\nu}) := \sum_{k=1}^q \nu_k \lambda_k.
\end{equation*}
The null signature $\mathbf{0}$ serves as the minimal element, while the full signature $\mathbf{m} = (m_1, \dots, m_q)$ acts as the maximal element.
The restriction to signatures of size exactly $r$, denoted $\cL_r:=\Set{\boldsymbol{\nu}\given \nu_1+\dotsb+\nu_q=r}$,
preserves this hierarchy and characterizes the critical projectors in $\cP_r$.
Each set $\cL_r$ contains two elements $\boldsymbol{\nu}^{\wedge,r}$ and $\boldsymbol{\nu}^{\vee,r}$
that correspond respectively to the minimal element, that saturates the selections of the smallest eigenvalues (including zeros), 
and to the maximal element, that saturate those of the largest eigenvalues.

\begin{figure}
	\centering
	\begin{tikzpicture}
		\node[minimum width=2cm,align=center] (L) {$Z\in\cT(\Id), \ \rk Z \le r$\\ cost $\fF$};
		\node[minimum width=2cm,align=center] (M) [right=2cm of L] {$P\in\cP_r$\\ cost $\fh$};
		\node[minimum width=2cm,align=center] (R) [right=2cm of M] {$U\in\cSt_r$\\ cost $\fk$};
		
		\draw[->, bend left=30] (L) to node[above] {$ZZ^+$} (M);
		\draw[->, bend left=30] (M) to node[below] {$A^{-1}Z_{\star}P$} (L);
		
		\draw[->, bend left=30] (M) to node[above,minimum width=1cm,align=center] {orthonormal\\ basis of $\Range P$} (R);
		\draw[->, bend left=30] (R) to node[below] {$UU^\top$} (M);
	\end{tikzpicture}
	\caption{\label{fig:thalweg} Geometric correspondence between the restricted critical thalweg, orthogonal projectors, and the Stiefel manifold.}
\end{figure}
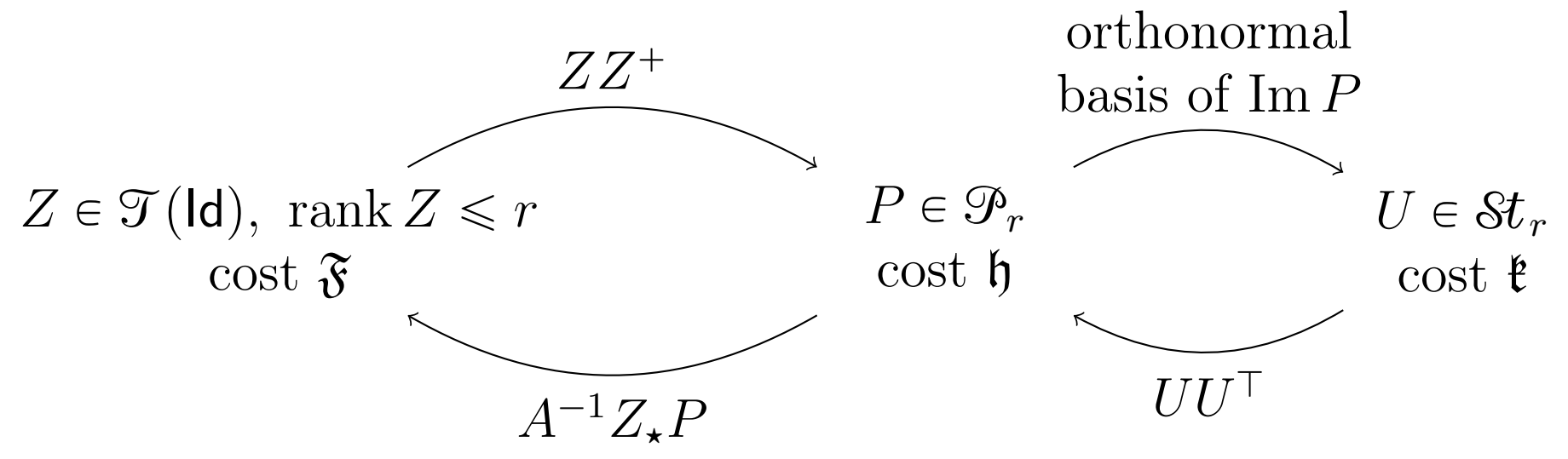

Denoting by $\cE_k$ the $m_k$-dimensional eigenspace associated with the eigenvalue $\lambda_k$, each signature $\boldsymbol{\nu} \in \cL_r$ of rank $0 \leq r \leq \dout$ defines a subset of projectors:
\begin{equation*}
    \cP({\boldsymbol{\nu}}) := \Set*{P\in\cP_r\given \Range(P)\subset \bigcup_{k=1,\dotsc,q,\ \nu_k>0} \cE_k \quad \text{with} \quad \rk(P_{|\cE_k}) = \nu_k}.
\end{equation*}
These signature-based sets partition the set of all rank-$r$ orthogonal projectors that commute with $\Sigma$.
Factorizing any $P \in \cP_r$ as $P = UU^\top$, where the columns of~$U$ form an orthonormal basis for its range, leads to the Stiefel manifold and the energy $\fk(U)$, defined for $1 \leq r \leq \dout$ by
\begin{equation*}
    \cSt_r := \Set{U\in\cM(\dout,r)\given U^\top U=\Id_r} \quad \text{and} \quad \fk(U) := \Tr(U^\top \Sigma U).
\end{equation*}
The cyclic property of the trace provides
\begin{equation*}
    \fh(P) = \Frob{\Sigma,P} = \Tr(\Sigma P) = \Tr(\Sigma U U^\top) = \Tr(U^\top \Sigma U) = \fk(U).
\end{equation*}
Applying a rotation to $U$ via any orthogonal matrix $R \in \cM(r,r)$ leaves $\fk(U)$ invariant:
\begin{equation} \label{eq:fk:invariance}
    \fk(UR) = \Tr((UR)^\top \Sigma (UR)) = \Tr(R^\top U^\top \Sigma U R) = \Tr(U^\top \Sigma U R R^\top) = \fk(U).
\end{equation}
Consequently, the energy $\fk$ depends solely on the column space of $U$. 
This interplay between the restricted critical matrices $\cT(\Id)\cap\cM_{r}(\din,\dout)$, the set of projectors~$\cP_r$, and the Stiefel manifold $\cSt_r$ establishes the geometric framework summarized in Figure~\ref{fig:thalweg}.

Building upon this framework, Theorem~\ref{thm:optimal_projector} resolves the problem of identifying critical points of $\fh$ over the subset $\cP_r$ via the spectral properties of $\Sigma$. 
It establishes an equivalence: a projector acts as a critical point if and only if it commutes with~$\Sigma$, which inherently restricts it to one of the signature-based classes $\cP(\boldsymbol{\nu})$ for some~$\boldsymbol{\nu} \in \cL_r$. 
For any such critical projector, the continuous energy evaluation reduces perfectly to the discrete level $\fh(P) = \fE(\boldsymbol{\nu})$, and the global maximum of $\fh$ over $\cP_r$ is therefore attained by the projectors belonging to the maximal class~$\cP(\boldsymbol{\nu}^{\vee,r})$.

As a byproduct, the proof of Theorem~\ref{thm:optimal_projector} recovers the celebrated Ky Fan theorem~\cite{Fan1949} and its min-max extension by Wielandt~\cite{Wielandt1955}. 
The extrema of $\fk$ are attained at matrices $U \in \cSt_r$ such that $UU^\top$ belongs to the extremal classes~$\cP(\boldsymbol{\nu}^{\vee,r})$ and~$\cP(\boldsymbol{\nu}^{\wedge,r})$. 
The critical points of $\fk$ are precisely the matrices $U \in \cSt_r$ satisfying $UU^\top \in \cP(\boldsymbol{\nu})$ for some $\boldsymbol{\nu}\in\cL_r$, with corresponding critical values $\fE(\boldsymbol{\nu})$.
Translating this characterization back to the original cost function $\fF$ yields Corollary~\ref{cor:rank_minimizer}, which explicitly provides the critical points and global minimizers of the rank-constrained loss.

\begin{theorem}
    \label{thm:optimal_projector}
    For any $P \in \cP_r$, the following condition are equivalent:
    \begin{enumerate}[thm]
        \item $P$ is a critical point of $\mathfrak{h}$.
        \item $P$ commutes with $\Sigma$.
        \item $P\in \cP(\boldsymbol{\nu})$ for some $\boldsymbol{\nu}\in\cL_r$, yielding the critical value $\fh(P) = \fE(\boldsymbol{\nu})$.
    \end{enumerate}
	Moreover, the global maximum of $\mathfrak{h}$ over $\cP_r$ is attained by the projectors in $\cP(\boldsymbol{\nu}^{\vee,r})$.
\end{theorem}

\begin{proof}
    From Equation \eqref{eq:fk:invariance}, the critical points of $\fk$ in $\cSt_r$ correspond to the critical points of $\fh$ on $\cP_r$ \cite{Edelman1998}. 
    The Riemannian gradient of $\mathfrak{k}$ with respect to $U$ under this constraint $U^\top U$ is 
    \begin{equation*}
        \nabla \fk(U) = 2 \left(\Sigma U - U (U^\top \Sigma U)\right).
    \end{equation*}
    Equating this gradient to zero and right-multiplying by $U^\top$ gives $\Sigma P = P \Sigma P$, which, by symmetry of $P$ and $\Sigma$, implies
    \begin{equation}
	\label{eq:P:Sig}
        P \Sigma = (\Sigma P)^\top = (P \Sigma P)^\top = P \Sigma P = \Sigma P.
    \end{equation}
    Therefore, $P$ is a critical point of $\fh$ if and only if $P$ and $\Sigma$ commute, which by simultaneous diagonalization \cite[Theorem~21.13.1, p.~568]{Harville2008a} is equivalent to $P\in\cP(\boldsymbol{\nu})$ for some index vector $\boldsymbol{\nu}$.

For $P\in\cP(\boldsymbol{\nu})$, simultaneous diagonalization provides an orthogonal matrix $W$ such that $W^\top \Sigma W = \Lambda$ and $W^\top P W = J$. 
Here, $\Lambda$ is the diagonal matrix containing the eigenvalues $\lambda_1>\dotsb>\lambda_q$ of $\Sigma$ with their multiplicities, and $J$ is a block-diagonal matrix whose $k$-th block of size $m_k$ has exactly $\nu_k$ ones and $m_k-\nu_k$ zeros on its diagonal. 
Consequently, injecting $P = W J W^\top$ and $\Sigma = W \Lambda W^\top$,
\begin{equation*}
    \fh(P) = \Tr(P \Sigma) = \Tr(W J W^\top W \Lambda W^\top) = \Tr(J \Lambda) = \sum_{k=1}^q \nu_k \lambda_k = \fE(\boldsymbol{\nu}).
\end{equation*}
Maximizing this sum over all valid index vectors yields the global maximum on the set $\cP(\boldsymbol{\nu}^{\vee,r})$.

\end{proof}

\begin{corollary}
    \label{cor:rank_minimizer}
    Any critical point of the loss $\fF$ restricted to $\cM_r(\din,\dout)$ 
    can be written as $Z = Z_\star P$ with $P\in\cP(\boldsymbol{\nu})$ for some $\boldsymbol{\nu}\in\cL_r$. In addition, 
    $P$ is the orthogonal projector onto the rows space of $Z$ and commutes with $\Sigma$. Finally,
    \begin{equation} \label{eq:value:F:critical}
	\mathfrak{F}(Z) = \frac{1}{2}\normF{Y}^2 - \frac{1}{2} \fE(\boldsymbol{\nu}).
    \end{equation}
    The global minimizers of the rank-constrained loss are thus matrices of the form $Z_r = Z_\star P_r$ with $P_r\in\cP(\boldsymbol{\nu}^{\vee,r})$.
\end{corollary}

\begin{proof}
If $Z$ is a critical point of $\fF$ restricted to $\cM_r(\din,\dout)$, 
then $Z\in\cT(\Id)$ and therefore $Z=Z_{\star}P$ for $P\in\cP_r$.
For any direction $\dot{P}$ in the tangent space to $\cP_r$ at point $P$, 
\begin{equation*}
    \dD\fh(P)\cdot \dot{P}
    =
    \dD\fF(Z_{\star}P)\cdot Z_{\star}\dot{P} 
    =
    \dD\fF(Z)\cdot Z_{\star}\dot{P} =0
\end{equation*}
by the very definition of a critical point. Thus, $P$ is a critical point of $\fh$
and $P\in\cP(\boldsymbol{\nu})$ for some $\nu\in\cL_r$. This leads to \eqref{eq:value:F:critical}
and to the characterization of $P$ by using Theorem~\ref{thm:optimal_projector} and Theorem~\ref{thm:thalweg_characterization}.
\end{proof}

%%%%%%%%%%%%%%%%%%%%%%%%%%%%%%%%%%%%%%%%%%%%%%%%%%%%%%%%%%%%%%%%%%%%%%

\subsection{Minima of the Rank-Constrained Problem}

The main result of this article, Theorem~\ref{thm:critical_points}, states
the absence of spurious local minima under the sole assumption that $X$ has
full column rank: whenever $\fG$ is non-constant, it admits neither local nor
global maxima, and every critical point is either a global minimum or a saddle
point connected to a global minimum by a continuous path along which~$\fG$ is
decreasing. Establishing this characterization proceeds in three stages. 
Proposition~\ref{prop:energy_ascent} operates on the set of fixed-rank
orthogonal projectors~$\cP_r$, constructing explicit continuous paths along
which~$\fh$ strictly increases from any critical
projector to a rank-$r$ global maximizer. Two intermediate lemmas connect the
projector geometry to the parameter space: Lemma~\ref{lem:no_maxima} rules out
local and global maxima of~$\fG$ via algebraic perturbations, while
Lemma~\ref{lem:crit} establishes a two-way correspondence between the
critical points of~$\fh$ and those of~$\fG$. The proof of
Theorem~\ref{thm:critical_points} combines the fixed-rank trajectories of
Proposition~\ref{prop:energy_ascent} with explicit rank-increasing paths
linking critical projectors in~$\cP_q$ to~$\cP_r$ ($q < r$), and lifts 
them as decreasing paths for~$\fG$ .

\begin{proposition} 
    \label{prop:energy_ascent}
	If $\boldsymbol{\nu} \prec \boldsymbol{\gamma}$ in $\cL_r$, any critical projector $P_{\boldsymbol{\nu}}\in\cP(\boldsymbol{\nu})$ 
	connects to a critical projector $P_{\boldsymbol{\gamma}}\in\cP(\boldsymbol{\gamma})$ via a continuous path in~$\cP_r$ strictly ascending the cost~$\fh$.
	In short, a strictly ascending path exists between any critical projector in~$\cP_r$ and the global optimum in~$\cP_r$.
\end{proposition}

\begin{proof}
	By Theorem~\ref{thm:optimal_projector}, two rank-$r$ critical projectors $P_{\boldsymbol{\nu}}$ and $P_{\boldsymbol{\gamma}}$ associated with signatures $\boldsymbol{\nu} \prec \boldsymbol{\gamma}$ take the forms $U_{\boldsymbol{\nu}} U_{\boldsymbol{\nu}}^\top$ and $U_{\boldsymbol{\gamma}} U_{\boldsymbol{\gamma}}^\top$, where $U_{\boldsymbol{\nu}}$ and $U_{\boldsymbol{\gamma}}$ are respectively formed by $r$ orthonormal eigenvectors $\mathsf{v}_1, \dots, \mathsf{v}_r$ and $\mathsf{w}_1, \dots, \mathsf{w}_r$ drawn from potentially distinct orthonormal eigenbasis of $\Sigma$. The indices can be chosen to mutually align the shared eigenvectors and pair the remaining ones by eigenvalue order, and the eigenvectors can be chosen up to a sign change such that $\mathsf{v}_k^\top \mathsf{w}_k \geq 0$.
	Denote by $\cK \subseteq \{1, \dots, r\}$ the set of indices $k$ where $\mathsf{v}_k$ and $\mathsf{w}_k$ are not collinear. Then, for each $k \in \cK$, define the targeted angle $\alpha_k = \arccos(\mathsf{v}_k^\top \mathsf{w}_k) \in (0, \pi/2]$ and the normalized projection $\mathsf{u}_k$ of $\mathsf{w}_k$ orthogonal to $\mathsf{v}_k$ as
	\begin{equation*}
	    \mathsf{u}_k = \frac{\mathsf{w}_k - (\mathsf{v}_k^\top \mathsf{w}_k)\mathsf{v}_k}{\norm{\mathsf{w}_k - (\mathsf{v}_k^\top \mathsf{w}_k)\mathsf{v}_k}}.
	\end{equation*}
	Setting $\alpha_{\max} = \max_{k \in \cK} \alpha_k$ and $\rho_k(\theta) = \theta \mathbf{1}_{\{\theta \leq \alpha_k\}} + \alpha_k \mathbf{1}_{\{\theta > \alpha_k\}}$, the rotation matrix
	\begin{equation*}
	    \sR(\theta) = \Id_{\dout} + \sum_{k \in \cK} \Big[ \sin(\rho_k(\theta))(\mathsf{u}_k\mathsf{v}_k^\top - \mathsf{v}_k\mathsf{u}_k^\top) + (\cos(\rho_k(\theta)) - 1)(\mathsf{v}_k\mathsf{v}_k^\top + \mathsf{u}_k\mathsf{u}_k^\top) \Big]
	\end{equation*}
	defines a continuous path $\sP(\theta) = \sR(\theta) P_{\boldsymbol{\nu}} \sR(\theta)^\top$ from $P_{\boldsymbol{\nu}}$ to $P_{\boldsymbol{\gamma}}$ over $[0, \alpha_{\max}]$, along which the cost function decouples into stationary and rotating components:
	\begin{equation*}
	    \fh(\sP(\theta)) = \sum_{k \notin \cK} \mathsf{v}_k^\top \Sigma \mathsf{v}_k + \sum_{k \in \cK} (\sR(\theta)\mathsf{v}_k)^\top \Sigma (\sR(\theta)\mathsf{v}_k).
	\end{equation*}
	Injecting $\sR(\theta)\mathsf{v}_k = \cos(\rho_k(\theta))\mathsf{v}_k + \sin(\rho_k(\theta))\mathsf{u}_k$ into the rotating components, the cross-terms $\mathsf{u}_k^\top \Sigma \mathsf{v}_k$ vanish by orthogonality. 
	The remaining terms evaluate exactly to $\mathsf{v}_k^\top \Sigma \mathsf{v}_k = \lambda_{v_k}$ and $\mathsf{u}_k^\top \Sigma \mathsf{u}_k = \lambda_{w_k}$, where $\lambda_{v_k}$ and $\lambda_{w_k}$ denote the respective eigenvalues of $\mathsf{v}_k$ and $\mathsf{w}_k$. The cost function thus simplifies to
	The cost function thus simplifies to
	\begin{align*}
	    \fh(\sP(\theta)) & = \sum_{k \notin \cK} \lambda_{v_k} + \sum_{k \in \cK} \left[ \cos^2(\rho_k(\theta)) \lambda_{v_k} + \sin^2(\rho_k(\theta)) \lambda_{w_k} \right] \\
	    & = \sum_{k=1}^r \lambda_{v_k} + \sum_{k \in \cK} \sin^2(\rho_k(\theta)) \left( \lambda_{w_k} - \lambda_{v_k} \right).
	\end{align*}
	Since $\boldsymbol{\nu} \prec \boldsymbol{\gamma}$, the inequality $\lambda_{w_k} \geq \lambda_{v_k}$ holds for all $k \in \cK$, with strict inequality for at least one index. 
	By construction $\alpha_k \in (0, \pi/2]$, meaning each function $\theta \mapsto \sin^2(\rho_k(\theta))$ is non-decreasing on $[0, \alpha_{\max}]$ and strictly increasing on $(0, \alpha_k)$. 
	Consequently, as $\theta$ progresses, the strictly positive differences ($\lambda_{w_k} > \lambda_{v_k}$) ensure the sum strictly increases on their respective intervals, rendering the overall cost function monotonically ascending along the entire path.
\end{proof}

\begin{lemma}
    \label{lem:no_maxima}
    The function $\mathfrak{G}$ has no local nor global maximum unless $\mathfrak{G}$ is constant.
\end{lemma}

\begin{proof}
	Since $\fG$ is non-constant, every layer dimension $d_0, \dotsc, d_p$ is positive, as a zero-dimensional layer would force every weight product to vanish and $\fG$ to be identically equal to $\fF(0)$. 
	Consider any tuple of weights $(V_1,\dotsc,V_p) \in \cM(d_0,\dotsc,d_p)$ and set $Z := \Psi_p(V_1,\dotsc,V_p) = V_1\dotsb V_p$.
	
	If $Z \neq 0$, scaling all weights by $\lambda > 0$ gives $\Psi_p(\lambda V_1,\dotsc,\lambda V_p) = \lambda^p Z$. 
	Expanding Equation~\eqref{eq:def:F} yields
	\begin{equation*}
	    \fG(\lambda V_1,\dotsc,\lambda V_p) = \fF(\lambda^p Z) = \frac{1}{2}\normF{Y}^2 - \lambda^p \Frob{Y, XZ} + \frac{\lambda^{2p}}{2}\normF{XZ}^2.
	\end{equation*}
	Because $X$ has full column rank and $Z \neq 0$, the last term is strictly positive. 
	Viewed as a function of $\mu := \lambda^p > 0$, the mapping $\mu \mapsto \fF(\mu Z)$ is a strictly convex quadratic polynomial diverging to $+\infty$ as $\mu \to +\infty$. 
	Since $\lambda \mapsto \lambda^p$ is a strictly increasing bijection on $(0, +\infty)$, the function $\lambda \mapsto \fG(\lambda V_1,\dotsc,\lambda V_p)$ admits no local maximum at $\lambda = 1$ and diverges to $+\infty$ as $\lambda \to +\infty$, ruling out any local or global maximum at $(V_1,\dotsc,V_p)$.
	
If $Z = 0$, then $\fG(V_1,\dotsc,V_p) = \fF(0)$. 
Set $A_0 := \Id_{d_0}$ and $m_0 := 0$, and define inductively for $k = 1, \dotsc, p$
\begin{equation*}
    (\widetilde{V}_k, A_k, m_k) := 
    \begin{cases}
        (V_k, A_{k-1}V_k, m_{k-1}) & \text{if } A_{k-1}V_k \neq 0, \\[4pt]
        (V_k + \varepsilon W_k, A_{k-1}W_k, m_{k-1} + 1) & \text{if } A_{k-1}V_k = 0.
    \end{cases}
\end{equation*}
Whenever $A_{k-1}V_k = 0$, since $A_{k-1} \neq 0$ by induction, there exists an index $j_k \in \{1,\dotsc,d_{k-1}\}$ such that the $j_k$-th column of $A_{k-1}$ is non-zero. 
Setting $W_k := e_{j_k} e_{d_k}^\top$, where $e_{j_k} \in \RR^{d_{k-1}}$ is the $j_k$-th canonical basis vector and $e_{d_k} \in \RR^{d_k}$ denotes the first canonical basis vector of $\RR^{d_k}$, yields $A_{k-1}W_k \neq 0$, so that by induction $\widetilde{V}_1 \dotsb \widetilde{V}_k = \varepsilon^{m_k} A_k$ with $A_k \neq 0$ for every $k$, while $Z = 0$ ensures $m_p \ge 1$.
With $Z' := A_p \neq 0$ and $\delta := \varepsilon^{m_p} > 0$, $\widetilde{V}_1 \dotsb \widetilde{V}_p = \delta Z'$, whereas reversing the sign of a single $W_k$ produces $-\delta Z'$.
Evaluating $\ff_{Z'}(\pm \delta) = \fF(\pm \delta Z')$ via Equation~\eqref{eq:f:1} yields
\begin{equation*}
    \ff_{Z'}(\pm \delta) = \fF(0) \mp \delta \Frob{Y, XZ'} + \frac{\delta^2}{2}\normF{XZ'}^2.
\end{equation*}
Since $\normF{XZ'}^2 > 0$, at least one of $\ff_{Z'}(\delta)$ or $\ff_{Z'}(-\delta)$ strictly exceeds $\fF(0)$ for sufficiently small $\delta > 0$, ruling out a local maximum at~$(V_1,\dotsc,V_p)$.
\end{proof}

\begin{lemma} 
    \label{lem:crit}
    Let $P$ be a critical point of $\fh$. Then there exists a critical point $(V_1,\dotsc,V_p)$ of $\fG$
    such that $V_1\dotsb V_p=Z_{\star}P$.
    Conversely, if $(V_1,\dotsc,V_p)$ is a critical point of $\fG$, then $V_1\dotsb V_p=Z_{\star}P$,
    where $P$ is an orthogonal projector that is a critical point of $\fh$.
\end{lemma}

\begin{proof}
    Since $P$ is a projector, there exists a $\dout\times r$-matrice $Q\in\cSt_r$ such that $P=QQ^\top$. 
    As $Z_{\star}=\Sigma_{X,X}^{-1}\Sigma_{X,Y}$,  
    \begin{equation*}
	- \nabla \fF(Z) = 
	\Sigma_{X,Y}-\Sigma_{X,X}Z_{\star}P=\Sigma_{X,Y}(\Id_{\dout}-P).
    \end{equation*}

    Using \eqref{eq:gradF:1}
    as well as the relationship between $\nabla \fF$ and $\nabla_{V_k}\fG$
    in the proof of Proposition~\ref{prop:critical_equivalence}, 
    $(V_1,\dotsc,V_p)=\Psi_p(Z_{\star}Q,Q)$ is a critical point of $\fG$ if and only if 
    \begin{equation*}
	\nabla \fF(Z) Q=0\text{ and }Q^\top Z_{\star}^\top \nabla \fF(Z)=0.
    \end{equation*}
    Since $Q^\top Q=\Id_r$, 
    \begin{equation*}
	-\nabla \fF(Z) Q = \Sigma_{X,Y}(\Id_{\dout}-P)Q = \Sigma_{X,Y}(\Id_{\dout}-QQ^\top)Q = 0.
    \end{equation*}
    This condition is always satisfied for any $Z=Z_{\star}QQ^\top$. This means
    that only $\nabla_{V_1}\fG(V_1,\dotsc,V_p)$ may not vanish.

    Regarding the second condition, as $Z_{\star}=\Sigma_{X,X}^{-1}\Sigma_{X,Y}$, 
    \begin{equation*}
	-Q^\top Z_{\star}^\top \nabla \fF(Z)
	=
	Q^\top \Sigma (\Id_{\dout}-P).
    \end{equation*}
    
    If $P$ and $\Sigma$ commute, since $Q^\top(\Id_{\dout}-P)=0$, $Z$ is a critical point for $\fF$ restricted to $\cM_{\rk(Z)}(\din,\dout)$, 
    $P$ is a critical point for $\fh$
    by Theorem~\ref{thm:optimal_projector} and $(V_1,\dotsc,V_p)$ is a critical point for $\fG$.

    Conversely, if $(V_1,\dotsc,V_p)$ is a critical point for $\fG$, then $Z=V_1\dotsb V_p$ belongs to the critical thalweg $\cT(\Id)$
    by Proposition~\ref{prop:critical_thalweg}
    and by Theorem~\ref{thm:thalweg_characterization}, there exists $P\in\cP_r$ with $r=\rk(Z)$ such that $Z=Z_{\star}P$
    There exists $U\in\cSt_r$ such that $P=UU^\top$.
    Since~$Z$ is a critical point of $\fF$ by Proposition~\ref{prop:critical_equivalence}, and $(Z_{\star}U,U^\top)$
    gives a rank factorization of $Z$, \eqref{eq:gradF:1} holds, 
    that is $Q^\top \Sigma (\Id_{\dout}-P)=0$ so that multiplying this equality on the left by $Q$, 
    $P\Sigma = P\Sigma P$. As in \eqref{eq:P:Sig}, $P$ and $\Sigma$ commute and 
    $P$ is a critical point for $\fh$ again by Theorem~\ref{thm:optimal_projector}.
\end{proof}

\begin{theorem}
    \label{thm:critical_points}
    Any critical point of $\fG$ is either a global minimum or is a saddle point that can be linked to the global 
    minimum with a path along which $\fG$ is decreasing. 
\end{theorem} 

\begin{proof}
    By Lemma~\ref{lem:no_maxima}, the function $\fG$ has no local nor global maxima unless $\fG$
    is constant in which case the theorem is trivial. 
    Therefore, $\fG$ is assumed to be non constant 
    so that any critical point of $\fG$ is either a saddle point or a minimum.
	Furthermore, Lemma~\ref{lem:crit} shows the equivalence between critical points of $\fG$
    and those of $\fF$. 
    The latter function is then considered first, by showing existence of paths between projectors having the same rank, and then different ranks.

    Let $P\in\cP_r$ be a critical point of $\fh$. 
    By Theorem~\ref{thm:optimal_projector}, $P\in\cP(\boldsymbol{\nu})$ for some $\boldsymbol{\nu}\in\cL_r$. 
    As $\boldsymbol{\nu}\prec \boldsymbol{\nu}^{\vee,r}$, Proposition~\ref{prop:energy_ascent} gives a continuous path $\sP:[0,1]\to\cP_r$ between $P$
    and any $P_r\in\cP(\boldsymbol{\nu}^{\vee,r})$ along which $\fh$ is increasing. Therefore, $\sZ:=Z_{\star}\sP$
    is a continuous path in $\cM_r(\din,\dout)$, as $\rk(\sZ(t))=\rk(\sP(t))=r$, with $\fF(\sZ(t))=\frac{1}{2}\normF{Y}^2  -\fh(\sP(t))$, 
    for $t\in[0,1]$. The cost function $\fF$ decreases along the path $\sZ$.

    Let $Z_0=Z_{\star}Q$ and $Q$ a critical point of~$\fh$ in~$\cP_q$, $q<r$. 
    Since $Q=UU^\top$, where $U$ is a $\dout\times q$-matrice
    whose columns are formed by $\Set{\sw_i}_{i\in\cK}$ where $\cK\subset\Set{1,\dotsc,\dout}$ with $\#\cK=q$ and the $\Set{\sw_i}_{i=1,\dotsc,\dout}$
    are the eigenvalues of $\Sigma$.

    Let $Z_0=Z_{\star}Q$ and $Q$ a critical point of~$\fh$ in~$\cP_q$, $q<r$. 
    Write $Q=UU^\top$, where the columns of $U \in \cM(\dout, q)$ are formed by $\Set{\sw_i}_{i\in\cK}$ for some subset $\cK\subset\Set{1,\dotsc,\dout}$ with $\#\cK=q$, and $\Set{\sw_i}_{i=1,\dotsc,\dout}$ are the eigenvalues of $\Sigma$.
    Then, choose $\cI$ to be a selection $r-q$ eigenvectors in $\Set{\sw_i}_{i=1,\dotsc,\dout}$ such that $\cI\cap\cK=\emptyset$. 
    Next, denote by~$V$ the matrix whose columns are the elements of $\Set{\sw_i}_{i\in\cI}$, and form 
    \begin{equation*}
    \sU(t) := \begin{bmatrix} 
	U & t V & \mathbf{0}_{\dout,\dout-r}
    \end{bmatrix}\text{ for } t\in[0,1]
    \end{equation*}
    so that 
    \begin{gather*}
    \sU(0) = \begin{bmatrix}
	U & \mathbf{0}_{\dout,\dout-q} 
    \end{bmatrix}
    ,\ 
    \sU(1) = \begin{bmatrix}
	U & V & \mathbf{0}_{\dout,\dout-r}
    \end{bmatrix}
    \\
    \text{and }\sU(t)\sU(t)^\top = UU^\top + t^2 VV^\top.
    \end{gather*}
    By Theorem~\ref{thm:optimal_projector}, $P=\sU(1)\sU(1)^\top\in\cP_r$ is a critical point of $\fh$. 
    Setting $\sZ(t):=Z_{\star}\sU(t)\sU(t)^\top$ defines a continuous path from the critical point $\sZ(0)=Z_0$ to the critical point~$\sZ(1)$, again according to Theorem~\ref{thm:optimal_projector}, even though $\sU(t)\sU(t)^\top$ is not a projector unless $t=0$ or $t=1$. 
    Combining~\eqref{eq:thalweg:7} with the inclusion $\sZ(t)\in\cS(\Id)$ for all $t\in[0,1]$ makes it sufficient to study
    \begin{equation*}
	    \fh(\sZ(t)) = \Tr\Paren*{\Sigma UU^\top} + t^2 \Tr\Paren*{\Sigma VV^\top}.
    \end{equation*}
    The symmetry of $\Sigma$ ensures that both $\Tr\Paren*{\Sigma UU^\top}$ and $\Tr\Paren*{\Sigma VV^\top}$ are non-negative, implying that $t\mapsto \fF(\sZ(t))$ is decreasing along $\sZ$, which connects a critical point of $\fF_{|\cM_q(\din,\dout)}$ to a critical point of $\fF_{|\cM_r(\din,\dout)}$.

    Let $(V_1,\dotsc,V_p)$ be some critical point of $\fG$. 
    As any critical point of $\fG$ corresponds to a critical point of $\fh$, 
    if there is no continuous, decreasing path for $\fG$ between $(V_1,\dotsc,V_p)$ and some 
    point at which $\fG$ reaches its global minimum, then there is no decreasing path for $\fF$ between $Z=V_1\dotsb V_p$ and a point $Z_{\dag}$ giving the global minimum. 
    As both $Z$ and $Z_{\dag}$ are respectively critical point of $\fF_{\cM_{|\rk(Z)(\din,\dout)}}$ and $\fF_{\cM_{|\underline{d}(\din,\dout)}}$, this contradicts the construction given above.
\end{proof}

%%%%%%%%%%%%%%%%%%%%%%%%%%%%%%%%%%%%%%%%%%%%%%%%%%%%%%%%%%%%%%%%%%%%%%
%%%%%%%%%%%%%%%%%%%%%%%%%%%%%%%%%%%%%%%%%%%%%%%%%%%%%%%%%%%%%%%%%%%%%%

\section{Global Shape of the Landscape and Numerical Experiments}

\label{sec:shape}

%%%%%%%%%%%%%%%%%%%%%%%%%%%%%%%%%%%%%%%%%%%%%%%%%%%%%%%%%%%%%%%%%%%%%%

\subsection{Landscape in the $(\rho, \theta)$-Plane}
This section provides a visualization tool which depends on the input and the output matrices~$X$ and~$Y$.
It offers a glimpse into the loss landscape, with a precise localization of the critical thalweg and therefore 
the critical points. 

With $\cV \coloneq \Set{\lambda \Id \given \lambda\in\RR}$,  
the generalized thalweg  $\cT_{\cV}(\Id)$ defined by \eqref{eq:thalweg-generalized}
is 
\begin{equation*}
    \cT_{\cV}(\lambda \Id) = \Set{Z\in\cM(\din,\dout) \given \Frob{Y,XZ} = \lambda \, \Frob{X Z, XZ }}
\end{equation*}
as $Z\in\cT_{\cV}(\lambda \Id)$ if and only if
\begin{equation}
    \label{eq:thalweg:6}
    \Frob{Y,XZ} = \lambda \,\Frob{XZ,XZ}.
\end{equation}

For $\lambda>0$, the ray restriction $\ff_Z(\lambda) := \fF(\lambda Z)$ expands as
\begin{equation}
    \label{eq:f:1}
    \ff_Z(\lambda) = \frac{1}{2}\normF{Y}^2 - \lambda \Frob{Y,XZ} + \frac{\lambda^2}{2}\normF{XZ}^2.
\end{equation}

Since $X$ has full column rank, for any $Z \neq 0$, the last term in Equation~\eqref{eq:f:1} is strictly positive, making $\ff_Z$ strictly convex. 
Its global minimizer is given by
\begin{equation*} 
    \alpha(Z) := 
    \begin{cases}
        \displaystyle \frac{\Frob{Y,XZ}}{\normF{XZ}^2} & \text{if } Z \neq 0, \\[10pt]
        0 & \text{otherwise.}
    \end{cases}
\end{equation*}
On the half-line $\RR_+$, $\ff_Z$ is increasing and attains its unique minimum at $\lambda = 0$ when $\Frob{Y,XZ} \le 0$.
When $\Frob{Y,XZ} > 0$, it attains its unique minimum at the interior point $\lambda = \alpha(Z) > 0$, decreasing on $[0, \alpha(Z)]$ and increasing on $[\alpha(Z), +\infty)$ from the minimal value
\begin{equation}
    \label{eq:f:2}
    \ff_Z(\alpha(Z)) = \frac{1}{2}\normF{Y}^2 - \frac{1}{2}\frac{\Frob{Y,XZ}^2}{\normF{XZ}^2}.
\end{equation}
For all $\lambda \in [0, 2\alpha(Z)]$, $\ff_Z(\lambda) \le \fF(0)$, with strict inequality on $(0, 2\alpha(Z))$, while $\ff_Z(\lambda) > \fF(0)$ for $\lambda > 2\alpha(Z)$. 
By Equation~\eqref{eq:thalweg:6}, for any $\lambda > 0$, $\cT_{\cV}(\lambda\Id)$ coincides with the level set $\alpha^{-1}(\{\lambda\})$, so $Z \in \cT_{\cV}(\lambda\Id)$ if and only if $\alpha(Z) = \lambda$, which forces $\Frob{Y,XZ} > 0$. Whatever the sign of $\Frob{Y,XZ}$, $\ff_Z$ and thus $\fF$ has no global maxima. This gives an alternative
proof to Lemma~\ref{lem:no_maxima}.

The landscape of the loss is analyzed through a reduced coordinate system for the end-to-end matrix $Z$ on the strip $\mathbb{R}_+ \times [0,\pi]$, referred to as the \emph{$(\rho,\theta)$-plane}. 
These coordinates are defined by extracting the relative scale $\rho(Z)$ and the angle $\theta(Z)$ between the prediction~$XZ$ and the target~$Y$:
\begin{equation*}
    \rho(Z) := \frac{\normF{XZ}}{\normF{Y}}
    \text{ and }
    \cos \theta(Z) := \frac{\Frob{Y,XZ}}{\normF{Y}\normF{XZ}}.
\end{equation*}

Recalling the expanded form of the loss $\mathfrak{F}(Z)$ from Equation~\eqref{eq:def:F}, dividing this expression by $\frac{1}{2}\normF{Y}^2$ and substituting the definitions of $\rho(Z)$ and $\cos\theta(Z)$ directly yields
\begin{equation*}
    \mathfrak{F}(Z) = \frac{1}{2} \normF{Y}^2 \, \mathfrak f(\rho(Z),\theta(Z)),
\end{equation*}
where the function $\mathfrak f : \mathbb{R}_+ \times [0,\pi] \to \mathbb{R}$ is defined as
\begin{equation*}
    \mathfrak f(\rho,\theta) := 1 - 2\rho \cos \theta + \rho^2.
\end{equation*}
Within the $(\rho,\theta)$-plane, $\mathfrak f(\rho,\theta)$ is strictly convex with respect to the scale $\rho$, but lacks both joint convexity in $(\rho, \theta)$ and global convexity with respect to the angle $\theta$.

The boundary and the upper half of the $(\rho,\theta)$-plane are characterized by elementary considerations:
\begin{enumerate}[noitemsep,topsep=-\parskip,label={(\roman*)}]
    \item If $Z=0$, then $\rho(Z)=0$, $\theta(Z)=\pi/2$, and $\fF(Z)=\fF(0)=\frac{1}{2}\normF{Y}^2$.
    \item If $\Frob{Y,XZ}=0$, then $\theta(Z)=\pi/2$, and $\fF(Z)=\frac{1}{2}\normF{Y}^2(1+\normF{XZ}^2)$.
	In the $(\rho,\theta)$-plane, the line $\RR_+\times\Set{\pi/2}$ corresponds to the horizontal line that cuts the strip in two parts.
    \item If $\Frob{Y,XZ}<0$, then $\theta(Z)>\pi/2$, and $\cF(Z)>\cF(0)$.
	The projection of $Z$ onto the $(\rho,\theta)$-plane belongs to the upper half-part of the strip.
\end{enumerate}

The focus shifts to the lower half of the strip, defined by the condition~$\Frob{Y,XZ}>0$.
For any $\mu>0$, the projection of $\cT_{\cV}(\mu)$ onto the $(\rho,\theta)$-plane yields the curve
\begin{equation*}
    \cU(\mu) := \Set*{(\rho,\theta) \in \mathbb{R}_+ \times \left[0,\frac{\pi}{2}\right) \given \cos \theta = \mu\rho}.
\end{equation*}

Since $Z$ belongs to $\cT_{\cV}(\alpha(Z))$, owing to \eqref{eq:f:2},
\begin{equation*}
    \mathfrak{F}(Z) 
	\begin{cases}
	    \le \mathfrak{F}(0) & \text{if }\alpha(Z) \in [0,2], \\[3pt]
	    = \fF(0) & \text{if }\alpha(Z)\in \Set{0,2},\\[3pt]
	    \ge \mathfrak{F}(0) & \text{if } \alpha(Z)\notin [0,2]
	\end{cases}
    \text{ and }
    \Set{\fF(Z)<\fF(0)}=\bigcup_{\mu\in(0,2)}\cT_{\cV}(\mu).
\end{equation*}
The set $\Set{\fF(Z)<\fF(0)}$ is  projected onto $\Set{(\rho,\theta)\given 0<\rho<2,\ 0<\theta<\arccos(2\rho)}$.

The $(\rho,\theta)$ representation is mostly useless in the scalar case as $\cos\theta(Z)\in\Set{-1,0,1}$.
In the general case, the projection depends on the input and output data $X$ and~$Y$.
When $\sP_X$ is the projector onto the columns of $X$, because $(\Id_{n}-\sP_X)^\top X=0$, 
\begin{equation*}
    \cos \theta(Z) = \frac{\Frob{\sP_X Y,XZ}}{\norm{\sP_X Y}\norm{XZ}}\frac{\norm{\sP_X Y}}{\norm{Y}}
    \in \left[-\frac{\norm{\sP_X Y}}{\norm{Y}},\frac{\norm{\sP_X Y}}{\norm{Y}}\right].
\end{equation*}
Even without any constraint on $Z$, $(\rho(Z),\theta(Z))$ can only reach a part 
of the strip~$\RR_+\times[0,\pi]$.
For the point $Z_\star$ given by \eqref{eq:value:Zstar}, 
\begin{equation*}
    \cos\theta(Z_\star)=\rho(Z_\star)=\frac{\norm{\sP_X Y}}{\norm{Y}}.
\end{equation*}

\begin{figure}[htbp]
    \centering
    \includegraphics[width=0.75\textwidth]{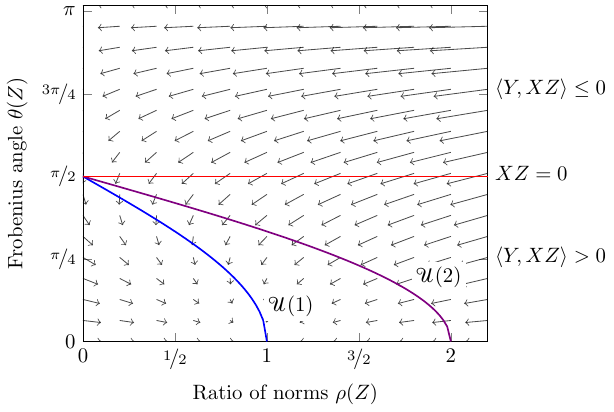}
    \caption{\label{fig:f}
    The negative gradient vector field $-\nabla \mathfrak{f}(\rho,\theta)$ in the polar plane. The blue curve represents the thalweg $\cU(1)$, where all critical points of $\mathfrak{G}$ lie. The horizontal red line at $\theta = \pi/2$ depicts the projection of matrices onto the kernel $\{Z \mid XZ = 0\}$. The violet curve marks the level set $\cU(2) = \mathfrak{f}^{-1}(\{1\})$, delineating the boundary where $\mathfrak{F}(Z) = \mathfrak{F}(0)$.}
\end{figure}

Figure~\ref{fig:f}, which displays the negative gradient vector field $-\nabla \ff(\rho,\theta)$
in the $(\rho,\theta)$-plane, synthesizes the main structural features of the loss
landscape. The blue curve representing the projection of the critical thalweg $\cU(1)$ attracts
the gradient flow. It contains the projection of the critical thalweg and that of all the critical points of~$\fG$.
Therefore, the critical thalweg acts as the primary valley floor.
Meanwhile, the violet level curve $\cU(2) = \fF^{-1}(\Set{0})$ delineates the
boundary separating regions of improved loss $\fF(Z) <
\fF(0)$ from degraded parameter configurations. Finally, the
horizontal red line at $\theta = \pi/2$ corresponds to the kernel subspace $\Set{Z
\given XZ = 0}$, along which the loss remains identically equal to its baseline
value $\mathfrak{F}(0)$.  We have empirically checked that the projection of
the gradient during the gradient descent remains rather close to $-\nabla \ff$.
Altogether, this polar representation hints how
the gradient dynamics are naturally channeled along the thalweg lines, guiding
the optimization process toward global minimizers.

%%%%%%%%%%%%%%%%%%%%%%%%%%%%%%%%%%%%%%%%%%%%%%%%%%%%%%%%%%%%%%%%%%%%%%

\subsection{The coefficient of determination}

The coefficient of determination for a matrix $Z$ of parameters
of the form $Z_\star P$ has not only a simple form but a clear geometric interpretation
in the $(\rho,\theta)$-plane.

We define 
\begin{equation*}
    \Rsq(Z) := \rho(Z)^2 
    \text{ so that }
    \Rsq(Z_\star)= \frac{\normF{XZ_{\star}}^2}{\normF{Y}^2}=\frac{\normF{\widehat{Y}}^2}{\normF{Y}^2}.
\end{equation*}
The coefficient $\Rsq(Z_{\star})$ is the \emph{raw coefficient of determination} --- or simply $\Rsq$ --- 
of the linear regression. It is one of the many metrics to quantify the quality of the regression, 
even among the ones called $\Rsq$ (see \textit{e.g.} \cite{Kvalseth,Luskin}).

From the properties of the projectors, $\normF{Y-\sP_X Y}^2 + \normF{\sP_X Y}^2 = \normF{Y}^2$.
Therefore, the value $\Rsq(Z_\star)$ is the \emph{uncentered coefficient of determination} and satisfies
\begin{equation*}
    \Rsq(Z_\star) = \Rsquncent(Z_{\star}):=1 -\frac{\normF{Y-\widehat{Y}}^2}{\normF{Y}^2}\leq 1.
\end{equation*}

If $\Frob{Y,\mathsf{1}_{n,\dout}}=0$, where $\mathsf{1}_{p,q}$ is the $(p,q)$-matrix whose coefficients
are equal to~$1$, then $\Rsq(Z_\star)$ is the proportion of explained variance, which corresponds
to the most classical definition of the coefficient of determination. 
The advantage of using $\Rsq$ over the quadratic loss function 
is that~$\Rsq$ does not depend on the norm of~$Y$ and may then be used to compare 
the behavior of the estimators over different sets of input and output data.

Proposition~\ref{prop:Rsq:formula} relates the accuracy of the prediction $Z_\star P\in\cT(\Id)$ is related
to the position of the projection of $Z_\star P$ onto the curve $\cU(1)$ in the $(\rho,\theta)$-plane.
For critical points, Corollary~\ref{cor:Rsq} the quality of the model is driven by the statistical distribution of the eigenvalues of~$\Sigma$.

\begin{proposition}[Formula for the $\Rsq$]
    \label{prop:Rsq:formula}
    Assume that $\Rsq(Z_\star)\neq 0$. For any $P\in\cP$, 
    \begin{equation}
	\label{eq:R2}
	\frac{\Rsq(Z_\star P)}{\Rsq(Z_\star)} = \frac{\Frob{\Sigma,P}}{\Frob{\Sigma,\Id_{\dout}}} 
	= \frac{\cos^2 \theta(Z_{\star}P)}{\cos^2 \theta(Z_{\star})}.
    \end{equation}
\end{proposition}

\begin{proof}
    For $P\in\cP$, as $P^2=P$ and $P^\top = P$, $P^\top P=P$.
    Since $XZ_{\star}=\sP_X Y$ and $\Sigma = (\sP_X Y)^\top \sP_X Y$, 
    \begin{equation*}
	\Frob{XZ_{\star} P ,XZ_{\star}P}
	=
	\Frob{\sP_X Y P, \sP_X Y P}
	=
	\Frob{\Sigma,P}.
    \end{equation*}
    Hence, 
    \begin{equation*}
	\Rsq(Z_\star P) = \frac{\Frob{\Sigma,P}}{\normF{\sP_X Y}^2}\frac{\normF{\sP_X Y}^2}{\normF{Y}^2}.
    \end{equation*}
    Noting that $\normF{\sP_X Y}^2=\Frob{\Sigma,\Id_{\dout}}$, we obtain \eqref{eq:R2}.
    Finally, since $\cZ(1)\subset\cT(1)$
    and $\cos\theta(Z)=\rho(Z)$ for any $Z\in\cT(1)$,  the relation involving $\theta(Z)$ and $\theta(Z_\star)$ holds.
\end{proof}

\begin{corollary}
    \label{cor:Rsq}
    For any $0\leq r\leq \dout$, $\boldsymbol{\nu}\in\cL_r$,
    and $P\in\cP(\boldsymbol{\nu})$, $P_r\in\cP(\boldsymbol{\nu}^{\vee,r})$
\begin{equation*}
    \frac{\Rsq(Z_{\star} P)}{\Rsq(Z_\star)} = 
	\frac{\fE(\boldsymbol{\nu})}{\fE(\boldsymbol{\mu}^{\vee,\dout})}
	%\frac{\sum_{k=1}^{q} \nu_k \lambda_k}{\sum_{k=1}^{q} m_k\lambda_k}
	\text{ and }
	\frac{\Rsq(Z_{\star} P)}{\Rsq(Z_{\star}P_r)} = 
    \frac{\fE(\boldsymbol{\nu})}{\fE(\boldsymbol{\nu}^{\vee,r})}.
\end{equation*}
\end{corollary}
%%

%%%%%%%%%%%%%%%%%%%%%%%%%%%%%%%%%%%%%%%%%%%%%%%%%%%%%%%%%%%%%%%%%%%%%%

\subsection{Numerical Experiments}

\label{sec:numerical}

The ability of steepest gradient descent to reach the critical
thalweg is assessed by sampling random pairs of matrices $(X,Y)$ and starting
points $(V,W)$, and running the algorithm from each starting point.

The metrics used to check if a pair $(V,W)$ is such that $VW$ is close to 
the generalized thalweg $\cT_{\RR\Id}(\Id)$ are 
(1) The distance 
\begin{equation*}
    d_{\text{gct}}:=\frac{1}{\din\times\dout}\abs{\Frob{\Sigma_{X,Y}-\Sigma_{X,X}VW,VW}}.
\end{equation*}
(2) The distance between $\rho$ and $\cos\theta$.
(3) The distance of the projection $VW$ to the $(\rho,\theta)$-plane 
given by $\delta:=(\rho-r)^2  + (\theta - \arccos(r))^2$
where $r$ solves  $\arccos(r) + \sqrt{1-r^2}(\rho - r) = \theta$.

The metric used to check if $(V,W)$ is such that $VW$ is close to 
the critical thalweg~$\cT(\Id)$ are 
(1) The distance 
\begin{equation*}
    d_{\text{ct}}:=\frac{1}{\din\times\dout}\normF{\Sigma_{X,Y}-\Sigma_{X,X}VW,VW}.
\end{equation*}
(2) For $P:=Z_\star^+ VW$, 
\begin{equation*}
    \delta_{\mathrm{sym}}:=\frac{\norm{P-P^\top}}{\dout^2}
\text{ and }
\delta_{\mathrm{idem}}:=\frac{\norm{P^2-P}}{\dout^2},
\end{equation*}
which are equal to $0$ when $P$ is a projector.
The normalized rank $r:=\Tr(P)/\dhid^2$ should
be close to $1$ when $VW$  is close to the critical thalweg.

The dimension are $\din=300$, $\dout=250$ and $\dhid=200$.
The input matrix $X$ is of size $300\times 250$ with random entries 
drawn according to the distributions Uniform, Gamma, Normal and Poisson. 
The output matrix is $Y=X+\epsilon$ for a perturbation~$\epsilon$ which is matrix 
with random entries $\epsilon\sim\mathcal{N}(0,1/10)$.

The gradient descent is run $\num{2000}$ steps. The mean is computed
over the last $20$ steps. This is performed \num{50} times and the metrics are averaged.
As shown in Table~\ref{table:1}, the last samples belong to the critical stream
and the loss continues to decrease, although at lower rate once
in the critical stream. The $\Rsq$ coefficients increases toward its maximal 
value. The norm of the gradient decreases, but never reaches $0$.
As already asserted theoretically, none of the critical points
excepted the minimal ones, have been reached. Although not shown here, 
 the gradient of~$\ff$ (plotted in Figure~\ref{fig:f})
seems to be a good proxy of the projection of the gradient of~$\fG$ during the gradient descent, 
even if not aligned.

\begin{table}[ht]
\begin{tabular}{r  c >{\footnotesize} c c >{\footnotesize} c }
    \toprule & \multicolumn{2}{c}{Uniform} & \multicolumn{2}{c}{Normal}  \\
 & Mean & Std & Mean & Std  \\
 \cmidrule(lr){2-3}
 \cmidrule(lr){4-5}
$\rho$ & \num{0.95} & \num{0.0058} &  \num{0.88} & \num{0.00069} \\
$\cos(\theta)$ & \num{0.95} & \num{0.0052} &  \num{0.88} & \num{0.0008} \\
$\theta$ & \num{0.31} & \num{0.017} &  \num{0.5} & \num{0.0017} \\
$d_{\text{ct}}$ & \num{0.00091} & \num{0.00024} &  \num{0.00077} & \num{0.00033} \\
$d_{\text{gct}}$ & \num{0.00019} & \num{0.00021} &  \num{0.00015} & \num{0.00011} \\
$\delta_{\mathrm{sym}}$ & \num{0.00039} & \num{2.2e-05} &  \num{5.7e-05} & \num{9.5e-06} \\
$\delta_{\mathrm{proj}}$ & \num{0.00013} & \num{4.4e-05} &  \num{1.7e-07} & \num{1.2e-07} \\
$r$ & \num{0.97} & \num{0.03} &  \num{1} & \num{0.00025} \\
$\delta$ & \num{4.8e-07} & \num{8.9e-07} &  \num{2.7e-08} & \num{3.9e-08} \\
$\Rsq$ & \num{0.91} & \num{0.011} &  \num{0.77} & \num{0.0012} \\
$\fG$ & \num{2.4e+03} & \num{2.6e+02} &  \num{1.8e+04} & \num{1.1e+02} \\
$\norm{\nabla V}$ & \num{42} & \num{12} &  \num{32} & \num{14} \\
$\norm{\nabla W}$ & \num{50} & \num{15} &  \num{27} & \num{12}  \\
\midrule & \multicolumn{2}{c}{Poisson} & \multicolumn{2}{c}{Gamma}  \\
 & Mean & Std & Mean & Std  \\
 \cmidrule(lr){2-3}
 \cmidrule(lr){4-5}
$\rho$ & \num{0.94} & \num{0.00027} &  \num{0.94} & \num{0.00022} \\
$\cos(\theta)$ & \num{0.94} & \num{0.00033} &  \num{0.94} & \num{0.00026} \\
$\theta$ & \num{0.34} & \num{0.00096} &  \num{0.34} & \num{0.00076} \\
$d_{\text{ct}}$ & \num{0.00078} & \num{0.00035} &  \num{0.00069} & \num{0.00029} \\
$d_{\text{gct}}$ & \num{0.00017} & \num{0.00012} &  \num{0.00013} & \num{8.8e-05} \\
$\delta_{\mathrm{sym}}$ & \num{5.9e-05} & \num{1e-05} &  \num{5.7e-05} & \num{8.6e-06} \\
$\delta_{\mathrm{proj}}$ & \num{1.6e-07} & \num{9.5e-08} &  \num{1.3e-07} & \num{6.6e-08} \\
$r$ & \num{1} & \num{0.00024} &  \num{1} & \num{0.00017} \\
$\delta$ & \num{7.7e-09} & \num{1.1e-08} &  \num{4.2e-09} & \num{5.7e-09} \\
$\Rsq$ & \num{0.89} & \num{0.00051} &  \num{0.89} & \num{0.00041} \\
$\fG$ & \num{1.8e+04} & \num{94} &  \num{1.8e+04} & \num{75} \\
$\norm{\nabla V}$ & \num{32} & \num{15} &  \num{28} & \num{12} \\
$\norm{\nabla W}$ & \num{22} & \num{12} &  \num{20} & \num{9.8}  \\
 \bottomrule
\end{tabular}
\caption{\label{table:1} Empirical mean of some statistics of the gradient descent on the last $20$ steps when $\num{2000}$ steps 
are performed in the gradient descent.}
\end{table}

\paragraph{\bfseries Acknowledgement.}
Saïd Toubra holds a Phd grant from CNRS.

%%%%%%%%%%%%%%%%%%%%%%%%%%%%%%%%%%%%%%%%%%%%%%%%%%%%%%%%%%%%%%%%%%%%%%
%%%%%%%%%%%%%%%%%%%%%%%%%%%%%%%%%%%%%%%%%%%%%%%%%%%%%%%%%%%%%%%%%%%%%%
%%%%%%%%%%%%%%%%%%%%%%%%%%%%%%%%%%%%%%%%%%%%%%%%%%%%%%%%%%%%%%%%%%%%%%
%%%%%%%%%%%%%%%%%%%%%%%%%%%%%%%%%%%%%%%%%%%%%%%%%%%%%%%%%%%%%%%%%%%%%%
%%%%%%%%%%%%%%%%%%%%%%%%%%%%%%%%%%%%%%%%%%%%%%%%%%%%%%%%%%%%%%%%%%%%%%
%%%%%%%%%%%%%%%%%%%%%%%%%%%%%%%%%%%%%%%%%%%%%%%%%%%%%%%%%%%%%%%%%%%%%%

\end{document}